\documentclass[10pt]{iopart}

\usepackage{cite}
\usepackage{iopams}
\usepackage{graphicx}
\usepackage{setspace}
\usepackage{epstopdf}
\usepackage{float}
\usepackage{trimclip}
\usepackage{caption}
\usepackage{subfiles}
\usepackage{footmisc}
\usepackage{subfigure}

\expandafter\let\csname equation*\endcsname\relax
\expandafter\let\csname endequation*\endcsname\relax
\usepackage{amsmath}

\usepackage[font=small]{caption}
\usepackage[export]{adjustbox}

\usepackage{makecell}
\usepackage[table]{xcolor}  
\usepackage{color, soul}
\makeatletter
\AtBeginDocument{\let\hl\@firstofone}
\makeatother

\usepackage{url}
\usepackage{algorithm}
\usepackage{etoolbox}\AtBeginEnvironment{algorithmic}{\scriptsize}
\usepackage[noend]{algpseudocode}
\usepackage{booktabs}
\usepackage{listings}
\usepackage{numprint}
\usepackage{siunitx}
\usepackage[nolist]{acronym}
\usepackage{breqn}
\usepackage[utf8]{inputenc}
\usepackage{todonotes}
\usepackage[shortcuts]{extdash}
\usepackage{textcomp}

\makeatletter
\newcommand\footnoteref[1]{\protected@xdef\@thefnmark{\ref{#1}}\@footnotemark}
\makeatother

\makeatletter
\def\lst@makecaption{%
  \def\@captype{table}%
  \@makecaption
}
\makeatother

\makeatletter
\def\BState{\State\hskip-\ALG@thistlm}
\makeatother

\usepackage[unicode]{hyperref}
\hypersetup{
  colorlinks,
  citecolor=black,
  urlcolor=black,
  linkcolor=black,
}

\hypersetup{
    pdftitle={Fast Collective Evasion in Self-Localized Swarms of Unmanned Aerial Vehicles},
    pdfpagemode=FullScreen,
    pdfauthor={Filip Novak, Viktor Walter, Pavel Petracek, Tomas Baca, Martin Saska},
}

\usepackage{tikz}
\usepackage{pgfplots}
\usepackage{filecontents}
\usetikzlibrary{shapes,fit,arrows,calc,intersections,positioning,backgrounds,decorations.markings,patterns,external}
\pgfplotsset{compat=newest}
\pgfdeclarelayer{background}
\pgfdeclarelayer{foreground}
\pgfsetlayers{background,main,foreground}

\tikzset{external/system call={latex \tikzexternalcheckshellescape -halt-on-error
    -interaction=batchmode -jobname "\image" "\texsource";
    dvips -o "\image".eps "\image".dvi;
ps2eps "\image.eps"}}

\tikzset{
  connect/.style args={(#1) to (#2) over (#3) by #4}{
    insert path={
      let \p1=($(#1)-(#3)$), \n1={veclen(\x1,\y1)},
      \n2={atan2(\x1,\y1)}, \n3={abs(#4)}, \n4={#4>0 ?180:-180}  in
      (#1) -- ($(#1)!\n1-\n3!(#3)$)
      arc (\n2:\n2+\n4:\n3) -- (#2)
    }
  },
}

\tikzset{
  state/.style={
    rectangle,
    draw=black, very thick,
    minimum height=1.0em,
    text centered,
  },
  final_state/.style={
    rectangle,
    rounded corners,
    draw=black, very thick,
    minimum height=2em,
    text centered,
  },
  initial_state/.style={
    rectangle,
    double=white,
    double distance=1pt,
    inner sep=2pt,
    draw=black, very thick,
    minimum height=2em,
    text centered,
  },
  point/.style={
    circle,
    inner sep=0pt,
    minimum size=3pt,
    fill=red
  },
  adder/.style={
    circle,
    inner sep=2pt,
    minimum size=0.3in,
    draw=black, very thick,
    text centered
  }
}

\usepackage{scalerel}
\usetikzlibrary{svg.path}
\definecolor{orcidlogocol}{HTML}{A6CE39}
\tikzset{
  orcidlogo/.pic={
    \fill[orcidlogocol] svg{M256,128c0,70.7-57.3,128-128,128C57.3,256,0,198.7,0,128C0,57.3,57.3,0,128,0C198.7,0,256,57.3,256,128z};
    \fill[white] svg{M86.3,186.2H70.9V79.1h15.4v48.4V186.2z}
    svg{M108.9,79.1h41.6c39.6,0,57,28.3,57,53.6c0,27.5-21.5,53.6-56.8,53.6h-41.8V79.1z M124.3,172.4h24.5c34.9,0,42.9-26.5,42.9-39.7c0-21.5-13.7-39.7-43.7-39.7h-23.7V172.4z}
    svg{M88.7,56.8c0,5.5-4.5,10.1-10.1,10.1c-5.6,0-10.1-4.6-10.1-10.1c0-5.6,4.5-10.1,10.1-10.1C84.2,46.7,88.7,51.3,88.7,56.8z};
  }
}
\newcommand\orcidicon[1]{\href{https://orcid.org/#1}{\mbox{\scalerel*{
        \begin{tikzpicture}[yscale=-1,transform shape]
          \pic{orcidlogo};
        \end{tikzpicture}
}{|}}}}

\begin{acronym}
  \acro{CNN}[CNN]{Convolutional Neural Network}
  \acro{MAV}[MAV]{Micro Aerial Vehicle}
  \acro{UAV}[UAV]{Unmanned Aerial Vehicle}
  \acro{UV}[UV]{ultraviolet}
  \acro{UVDAR}[UVDAR]{UltraViolet Direction and Ranging}
  \acro{GNSS}[GNSS]{Global Navigation Satellite System}
  \acro{RTK}[RTK]{Real-Time Kinematic}
  \acro{GPS}[GPS]{Global Positioning System}
  \acro{LiDAR}[LiDAR]{Light Detection and Ranging}
  \acro{RADAR}[RADAR]{Radio Detection and Ranging}
  \acro{LED}[LED]{Light-Emitting Diode}
  \acro{MPC}[MPC]{Model Predictive Control}
  \acro{AUV}[AUV]{Autonomous Underwater Vehicle}
\end{acronym}

\newcommand{\threeD}{\mbox{3-D}}
\newcommand{\twoD}{\mbox{2-D}}
\newcommand{\norm}[1]{||#1||}
\renewcommand{\vec}[1]{\boldsymbol{#1}}

\newcommand{\PREPRINTYEAR}{2021}
\newcommand{\PUBLISHEDIN}{ IOP Publishing}
\newcommand{\DOI}{10.1088/1748-3190/ac3060} 

\usepackage[placement=top,vshift=25,hshift=30]{background}
\SetBgScale{1.0}
\SetBgContents{PREPRINT VERSION - DO NOT DISTRIBUTE. PROPERTY OF IOP PUBLISHING. \href{https://doi.org/\DOI}{DOI \DOI}}
\SetBgColor{black}
\SetBgAngle{0}
\SetBgOpacity{1.0}

\begin{document}

\thispagestyle{empty}
\onecolumn
{
  \topskip0pt
  \vspace*{\fill}
  \centering
  \LARGE{%
    \copyright{} \PREPRINTYEAR~\PUBLISHEDIN\\\vspace{1cm}
    Personal use of this material is permitted.
    Permission from \PUBLISHEDIN~must be obtained for all other uses, in any current or future media, including reprinting or republishing this material for advertising or promotional purposes, creating new collective works, for resale or redistribution to servers or lists, or reuse of any copyrighted component of this work in other works.\vspace{1cm}\newline
    DOI: \href{https://doi.org/\DOI}{\DOI}}
    \vspace*{\fill}
}
\NoBgThispage
\BgThispage
\newpage

\title[Fast Collective Evasion in Self-Localized Swarms of Unmanned Aerial Vehicles]{Fast Collective Evasion in Self-Localized Swarms of Unmanned Aerial Vehicles}

\author{
  Filip Nov\'{a}k$^{1\orcidicon{0000-0003-3826-5904}}$
  Viktor Walter$^{\orcidicon{0000-0001-8693-6261}}$,
  Pavel Petr\'{a}\v{c}ek$^{\orcidicon{0000-0002-0887-9430}}$,
  Tom\'{a}\v{s} B\'{a}\v{c}a$^{\orcidicon{0000-0001-9649-8277}}$,
  Martin Saska$^{\orcidicon{0000-0001-7106-3816}}$
}

\address{Department of Cybernetics, Faculty of Electrical Engineering, Czech Technical University in Prague, 166 36, Prague 6, Czech Republic}
\address{$^1$Author to whom any correspondence should be addressed.}
\eads{\mailto{filip.novak@fel.cvut.cz}, \mailto{viktor.walter@fel.cvut.cz}, \mailto{pavel.petracek@fel.cvut.cz}, \mailto{tomas.baca@fel.cvut.cz}, \mailto{martin.saska@fel.cvut.cz}}
\vspace{10pt}
\begin{indented}
\item[]October 2021
\end{indented}

\begin{abstract}

A novel approach for achieving fast evasion in self-localized swarms of Unmanned Aerial Vehicles (UAVs) threatened by an intruding moving object is presented in this paper.  
Motivated by natural self-organizing systems, the presented approach of fast and collective evasion enables the UAV swarm to avoid dynamic objects (interferers) that are actively approaching the group. 
The main objective of the proposed technique is the fast and safe escape of the swarm from an interferer ~discovered in proximity.
This method is inspired by the collective behavior of groups of certain animals, such as schools of fish or flocks of birds.
These animals use the limited information of their sensing organs and decentralized control to achieve reliable and effective group motion.
The system presented in this paper is intended to execute the safe coordination of UAV swarms with a large number of agents.
Similar to natural swarms, this system propagates a fast shock of information about detected interferers throughout the group to achieve dynamic and collective evasion.
The proposed system is fully decentralized using only onboard sensors to mutually localize swarm agents and interferers, similar to how animals accomplish this behavior. 
As a result, the communication structure between swarm agents is not overwhelmed by information about the state (position and velocity) of each individual and it is reliable to communication dropouts.
The proposed system and theory were numerically evaluated and verified in real-world experiments. 
\end{abstract}

\vspace{2pc}

\submitto{\BB}

\maketitle
 
\ioptwocol

\section{Introduction}
Reliable collective motion of tightly cooperating aerial robots in real-world conditions is becoming required in numerous application scenarios.
The deployment of groups of cooperating \acp{UAV} is often motivated by the reduction of overall mission time, increased redundancy, or applying heterogeneous teams of \acp{UAV} that better replace an equipped single \ac{UAV} with a bigger payload.
A multi-\ac{UAV} system can be used to search for lost people over a wide area, for the localization of fires and initial fire-fighting actions, for measuring pollution over cities, taking measurements of radiation, monitoring animals, ensuring security and surveillance of large areas, and helping to ensure security at large events, such as festivals.
In all of these applications, a large number of small cameras or other sensors distributed in the work-space provided more information than one high quality camera.

\begin{figure}[!b]
\centering
\includegraphics[width=\columnwidth]{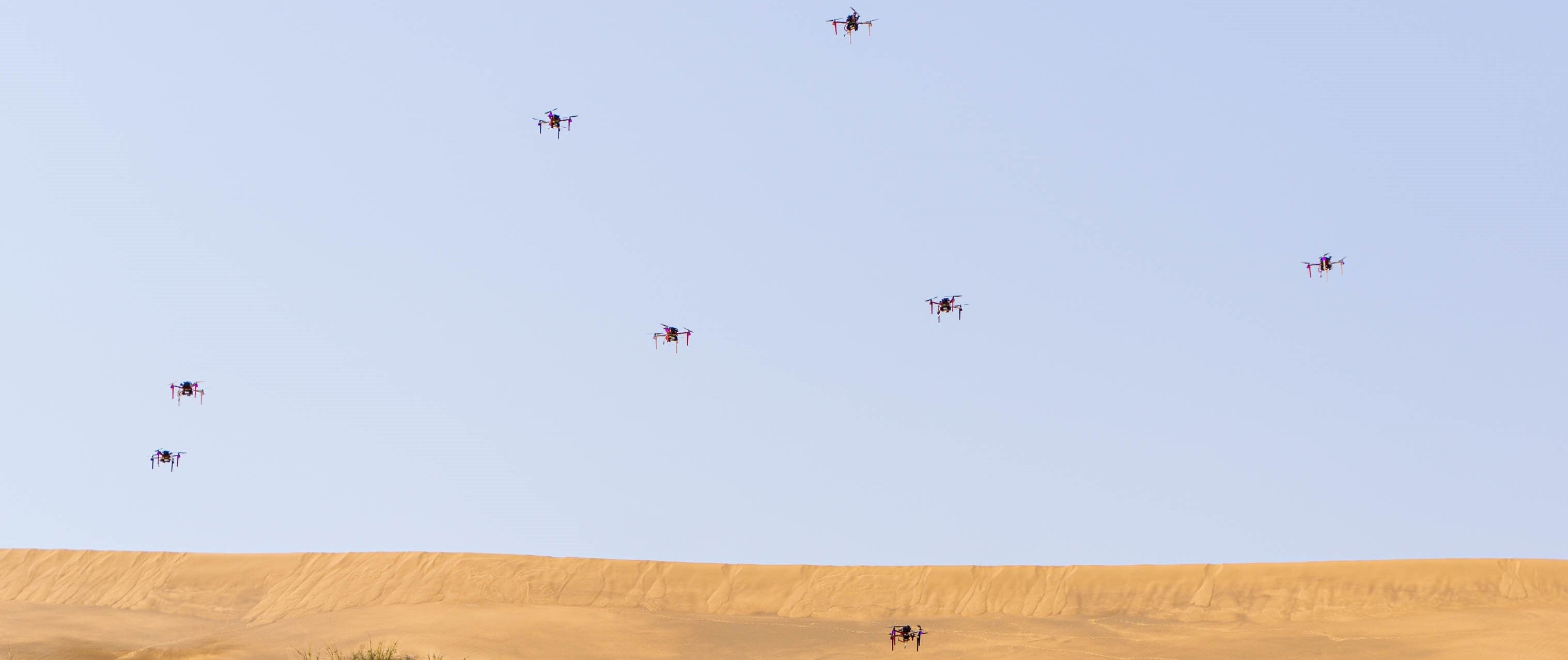}
\caption{A compact \ac{UAV} group stabilized above dunes in a desert using some of the swarming principles and visual relative localization being used in this paper.}
\label{fig:UAVgroup}
\end{figure}

\begin{figure}[!t]
\centering
\includegraphics[width=\columnwidth]{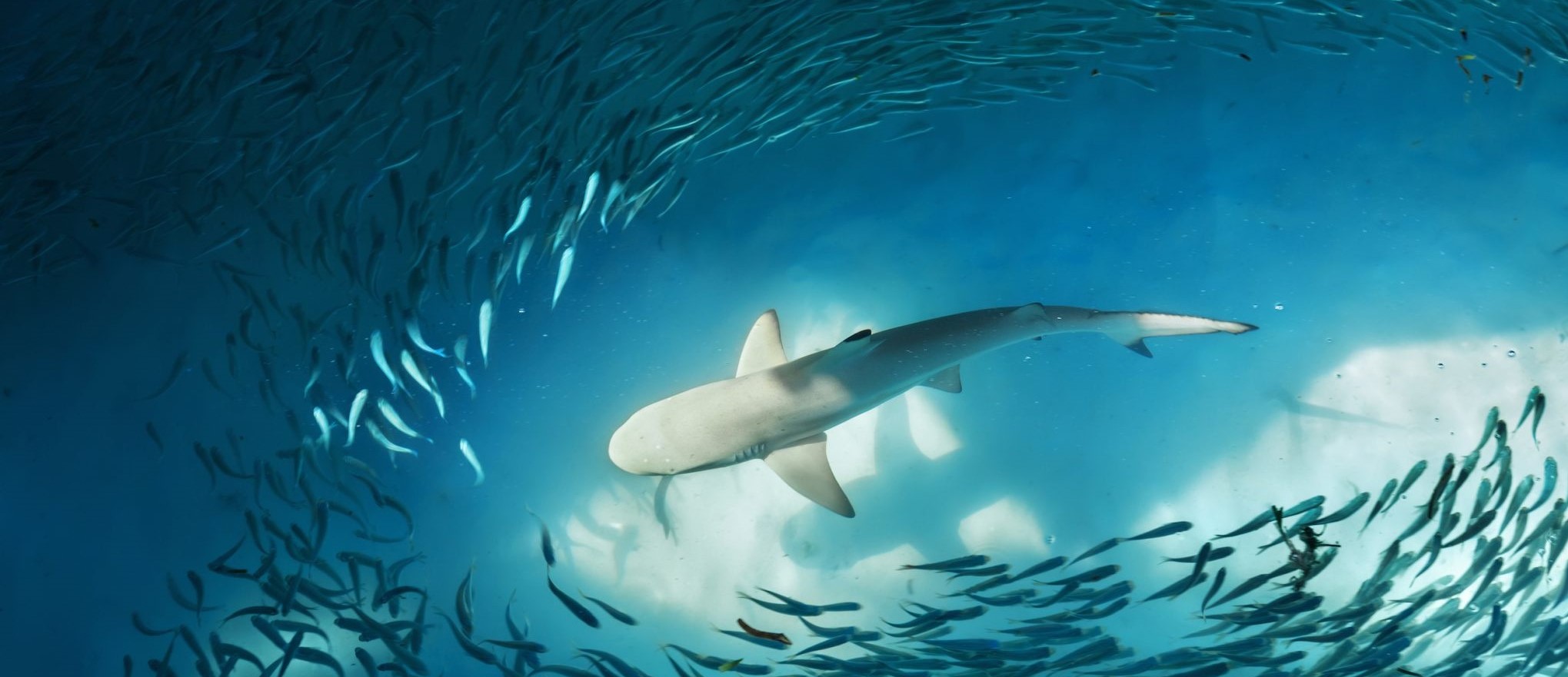}
\caption{Self-organizing social behavior protecting individuals in a school of fish from an interferer.}
\label{fig:shark_and_fish}
\end{figure}

Using a large group of closely cooperating \acp{UAV} (Figure~\ref{fig:UAVgroup}) in a real-world environment may require the ability to rapidly evade objects approaching the group.
The object can be viewed as a moving obstacle or an interferer in case of a continuous or apparently deliberate approach towards the \ac{UAV} group.
Numerous objects can be considered interferers, such as another aerial vehicle, people trying to negatively interact with the group of \acp{UAV} (e.g. playing children), hostile devices trying to assault the group, and also well known attacks on drones by animals. 
The evasion from such objects is called $evasive$ $behavior$. 
It can protect the group of \acp{UAV} and, often more importantly, the interfering object itself. 
For example, children coming close to a group of \acp{UAV} could potentially get hurt.

The evasive behavior in large compact groups, also referred to as \ac{UAV} swarms, can be seen in nature for animals in a defensive situation, e.g. a school of fish evading a shark, as shown in Figure \ref{fig:shark_and_fish}.
In such situations, it is common that only part of the school detects the interferer due to their orientation, distance, occultation, and other sensory limitations.
Information about the detected interferer is then spread throughout the entire school.
The dissemination of this information is called $shock$ $propagation$.
Such limited sensory capability is shared with \ac{UAV} swarms equipped with local onboard sensing. 
To be able to mimic the group behavior of animals in large \ac{UAV} swarms, various nature inspired swarm models can be applied \cite{swarmBook, Smith_2019, OH201783, Vir_gh_2014}.
These principles conform to requirements related to scalability, decentralized control, and lower dependency on communication, but the collective evasion was not studied in any of them.

In this paper, we propose a novel approach of evasive behavior applicable to most of the available swarm models. 
The~novel approach satisfies all the positive properties of nature-inspired swarming, such as scalability, fully decentralized control, exclusive reliance on onboard sensors, independence to any external infrastructure, and low communication requirements. 
It also significantly increases \ac{UAV} swarm safety and reliability in the event of dynamic obstacles and actively approaching interferers in particular. 
To demonstrate the general usability of the proposed approach, the classical Boids swarm model \cite{boids} is used for stabilization of the group for the experimental evaluation of the proposed mechanism.
However, any high-level multi-\ac{UAV} coordination can be used if information about neighboring \acp{UAV} is available.
To localize other swarm agents and to reduce the requirements for communication, the proposed swarm system uses an onboard relative localization system called \ac{UVDAR} \cite{UVDAR, UVDAR1} that we developed specifically for compact \ac{UAV} swarms.
The system presented in this paper was successfully verified in a realistic simulator and by conducting real-world experiments.
The \ac{UAV} swarm repeatedly avoided an interferer while the swarm members maintained a compact group and kept a safe distance between themselves.

\section{Related work}
Robotic swarming is the subject of research in many scientific studies \cite{swarmDef1, swarmDef2, Olfati, Berlingereabd8668, takeoffSwarm, Elamvazhuthi_2019}. 
One of the oldest models for swarm stabilization is the Boids system, introduced in~\cite{boids}.
At its core, this system was inspired by observing flocks of birds. 
It was designed for dimensionless particles in computer graphics, with the real-world dynamics of these particles not being considered.
In this work, we adapt the Boids system rules for the basic stabilization of the \ac{UAV} team and for experimental verification of the proposed collective evasion mechanism with a shock propagation.

The Boids model was also used as motivation in~\cite{Hauert2011}, which focuses on fixed-wing robots using network communication during flight and a \ac{GPS} module for localization. 
A swarm optimization algorithm inspired by the behavior of starling birds is proposed in \cite{FlockOpt}.
The principles in \cite{FlockOpt} are based on Boids rules.
However, the work does not deal with real-world deployment as evaluation of the system takes place in a simulator with dimensionless particles. 
Another study, \cite{Brkle2011}, presents a method developed for rotary-wing \acp{MAV}.
The \acp{MAV} were equipped with a network module for communication between them and a ground control station. 
Although all three works were evaluated in a simulator and works  \cite{Hauert2011, Brkle2011} in real-world experiments, neither static nor dynamic obstacles were considered and no evasive mechanism was studied.

Flocking of quad-rotor \acp{UAV} was presented in \cite{Vasarhelyi} where the method was successfully verified with real robots. 
However, the method in \cite{Vasarhelyi} still requires explicit sharing of global positions between the \acp{UAV}. 
These positions were obtained from onboard \ac{GNSS} receivers, making the system unsuitable for \ac{GNSS} denied environments, such as the indoors or forests.
Additionally, \ac{GNSS} precision can often be insufficient for \acp{UAV} flying in close proximity.
Like the works mentioned above, the method in \cite{Vasarhelyi} does not provide any obstacle avoidance.

A swarm of \acp{MAV} using an onboard relative localization system is presented in \cite{Nageli}. 
The relative localization is performed using a camera and artificial markers placed on each swarm member.
However, the system in \cite{Nageli} requires communication between swarm members, and obstacle avoidance is not included.
A~unique approach to swarming without communication was introduced in \cite{BioInspiredSwarm}. 
To localize neighboring agents, \acp{UAV} used the same onboard relative localization system \cite{UVDAR, UVDAR1} as used in this paper. 
Although static obstacle avoidance is integrated in this system, neither dynamic obstacles nor any evasive mechanism were part of the study in \cite{BioInspiredSwarm}.

Collision and obstacle avoidance in a leader-followers swarm model is presented in \cite{collisionAvoidanceLeaderFollower, Lee2018, collisionAvoidanceSwarm}. 
However, these approaches focus on static obstacle avoidance and broadcast full states. 
Static obstacle avoidance is also the focus of studies in \cite{Huang2021, obstacleAvoidanceAlgSwarm} that were evaluated by simulations.
The approach in \cite{obstacleAvoidanceAlgSwarm} was further verified in real-world experiments.
The study in \cite{TAYLOR2021103754} is focused on collision avoidance between swarm members and was evaluated by simulations.

Another approach to obstacle avoidance is with generation of collision-free paths for individual \acp{UAV}~\cite{ZHEN2020105826, obstacleAvoidanceManager, DCADcollisionAvoidance, Ashraf2020, navigationSwarmDrones}. 
However, these approaches are verified only in simulations and require sufficient localization in map frame to compute collision-free paths for individuals \acp{UAV}.
Collision avoidance between swarm members using communication and non-moving obstacle avoidance using \ac{LiDAR} is presented in \cite{obstacleAvoidanceManager}.   
Collision-free paths considering static as well as moving obstacles is proposed in \cite{Ashraf2020}, but precise states of all \acp{UAV} have to be shared over the swarm for the correct function of this approach. 

Studies based on the Chaotic Ant Colony Optimisation for Coverage (CACOC) model applied to a \ac{UAV} swarm are presented in \cite{Dentler2019, optPerformanceUAVSwarmIntruder}. 
The approach in \cite{Dentler2019} extends CACOC with a method of collision avoidance and was tested only in simulations while broadcasting the full states of all \acp{UAV}.
The study in \cite{optPerformanceUAVSwarmIntruder} proposes a method to increase intruder detection rate with a \ac{UAV} swarm based on the CACOC model.
This method is analyzed in simulations in which the states of \acp{UAV} are shared.

Similar characteristics and challenges in a \ac{UAV} swarm are detected in a swarm of \acp{AUV} \cite{CoCoRo}, as \acp{AUV} can move in any direction of \threeD{} space. 
The work in \cite{CoCoRo} provides a bio-inspired robotic system that is scalable, reliable, and flexible.
However, the system in \cite{CoCoRo} does not deal with the avoidance of dynamic obstacles actively approaching the swarm.
Moreover, the communication channels are used for interactions between individual swarm members.

An escape behavior mechanism intended to avoid an approaching object is proposed in \cite{MinWang, Min2010}. 
The authors presented a decentralized control algorithm applied for ground robots in simple \twoD{} space with usage of external global positioning.
The main focus of the work in \cite{MinWang, Min2010} uses observations of rapid change in the heading of neighboring robots for collective escape action, which is not suitable for \acp{UAV} as they can move in any direction without heading change. 
Among others, the consideration of high dynamic constraints of swarming \acp{UAV} and constraints given only by local sensing are our primary contributions in this paper.
Another approach of predator avoidance method is based on providing unpredictable trajectories \cite{Curiac_2015}. 
These random zig-zag movements have to confuse the predator while robots still perform their patrolling tasks.

A mathematical study of the interaction between an interferer and a herd of sheep is presented in~\cite{predatorSwarmInteraction}. 
The research focuses on the shape of the swarm interacting with the interferer, with the dynamics of the resulting model also being analyzed. 
Similar to the works in \cite{MinWang, Min2010}, the interferer-swarm interaction is studied in \twoD{} and none of the requirements of \ac{UAV} deployment in real environments are considered.

For using evasive behavior in the real world, interferer localization has to be implemented.
In real applications, an interferer will not cooperate by broadcasting its state (position) or carrying artificial visual markers.
Therefore, localization methods based on markers mounted on an interferer are not applicable \cite{markersLocKrajnik, markersLoc2, markersLoc3}.
The method in \cite{radarLoc} is based on a combination of \ac{RADAR} and \ac{LiDAR}, which is meant to be used for long-range detection using ground sensors. 
However, onboard relative localization in short ranges can not be obtained.
More suitable approaches include markerless methods based on computer vision using an onboard camera and a \ac{CNN}, as is the approach that we had developed originally for the interception of unauthorized drones \cite{visionVrba1, visionVrba2}.
These methods do not require external markers on target and are based on processing images from an onboard camera that is lightweight.
For this reason, these methods could be used for interferer localization. 
Thus, the proposed collective evasion approach is well suited for their usage. 
However, any other moving object localization can be used with the proposed swarming approach, as shown in the experimental section of this paper.

\begin{table*}
\centering
\begin{tabular}{|c|c|c|c|c|}
\hline
Work & Mutual localization & Designed for UAVs & \makecell{Dynamic obstacle\\avoidance} & \makecell{Real-world\\ deployment}\\
\hline
Burkle et al. \cite{Brkle2011} & Shared global position & Yes & No & Yes\\
Hauert et al. \cite{Hauert2011} & Shared global position & Yes & No & Yes\\
Vasarhelyi et al. \cite{Vasarhelyi} & Shared global position & Yes & No & Yes\\
Nageli et al. \cite{Nageli}  & Onboard relative localization & Yes & No & Yes\\
Petracek et al. \cite{BioInspiredSwarm} & Onboard relative localization & Yes & No & Yes\\
Ahmad et al. \cite{afzal2021Icra} & Onboard relative localization & Yes & No & Yes\\
Min \& Wang \cite{MinWang, Min2010} & Shared global position & No & Yes & Yes\\
Ashraf et al. \cite{Ashraf2020} & Shared global position & Yes & Yes & No\\
\textbf{This work} & \textbf{Onboard relative localization} & \textbf{Yes} & \textbf{Yes} & \textbf{Yes}\\
\hline
\end{tabular}
\caption{A brief comparison of swarm systems related to this work.}
\label{table:ComparisonRelatedWorks}
\end{table*}

\subsection{Contributions}
This paper addresses the challenges related to encounters of real swarm systems with dynamic obstacles.
One of the primary tasks of preventing collisions between swarm agents and dynamic obstacles is solved using a novel method for fast collective evasion.
To summarize, the main contributions of this paper include:
\begin{itemize}
	\item a method for fast collective evasion of multi-rotor \acp{UAV} to avoid dynamic obstacles designed and verified in the real world,
	\item the analysis of shock propagation using communication with limited bandwidth that allows for the use of implicit communication only,
	\item the designed swarming method using onboard relative localization system \ac{UVDAR} as was verified in multiple real-world experiments.
\end{itemize}

\subsection{Problem statement}
In this paper, we tackle the problem of the decentralized coordination of a team of closely flying small-size \acp{UAV} (less than 0,6 m diameter) without the use of explicit communication with any central station. 
We assume a team of homogeneous \acp{UAV} (referred to as swarm \acp{UAV} in this paper) equipped with a low-level stabilization relative to the environment. 
It means that each \ac{UAV} is able to take off at some place and be navigated into a given reference in its own reference frame by a position controller (such as we provided in \cite{baca2020mrs}).
In addition, each \ac{UAV} is equipped with an onboard localization system that is able to estimate the relative positions (distance and bearing) of all its teammates within a given range of 8 m (such as provided by the \ac{UVDAR} localization system \cite{UVDAR, UVDAR1}). 
Each \ac{UAV} should be able to communicate with others using a simple network to share messages including its timestamp, \ac{UAV} name, and bool variable determining presence of an interferer with a minimal rate of one message per second.
Furthermore, we assume that the localization system (or another onboard system) is able to provide an implicit communication link with a bandwidth 0.5 bps between all \acp{UAV} within the localization range. 
For example, such implicit communication with extremely low bandwidth can be done through observation of active markers of the \ac{UVDAR} system blinking with varying frequency. 
Thus, any classical wire-less communication is not needed.

We also assume the presence of a flying interferer that is actively pursuing the swarming \acp{UAV}. 
We assume that the interferer can fly with a maximum speed that is 90\% of the maximum speed of the swarming \acp{UAV} and has the same or tighter dynamic constraints of the swarming \acp{UAV}. 
We assume that each of the swarming \acp{UAV} can detect the interferer within a given distance of 12~m according to the size of \acp{UAV} used in this paper.
In the case of static obstacles in the environment, we assume that the \acp{UAV} are equipped with an onboard system for localization of these objects, such as the approach we proposed in \cite{kratky2021exploration} where the map is unknown and cannot be shared within the swarm. 
Although we have demonstrated the ability of the \ac{UAV} swarm to detect and avoid static obstacles~\cite{afzal2021Icra}, this paper is focused on the greater challenge of avoidance of dynamic obstacle actively approaching the group. 
Thus, static obstacles are not considered in the experimental section of this paper to clearly present the intended fast collective evasion.

\subsection{Comparison with related works}
A comparison of this work with the most related works is summarized in Table \ref{table:ComparisonRelatedWorks}.
Many of the works rely on sharing global position for mutual localization.
Therefore, a reliable communication structure with appropriate rate and bandwidth is required.
Our approach uses an onboard \ac{UVDAR} system with no communication requirement.
The swarm systems mentioned in the first six rows of the table do not provide any dynamic obstacle avoidance.
The system presented in \cite{MinWang, Min2010} proposes dynamic obstacle avoidance. However, the system is designed for ground robots moving in \twoD{} space and requires sharing the global positions via a reliable communication channel. 
This is difficult to achieve in real-world conditions and for larger groups.
The approach in \cite{Ashraf2020} is based on planning collision-free paths and is able to avoid dynamic obstacles.
However, the system requires broadcasting the precise \acp{UAV} states for swarm stabilization, and the system was not designed for a real-world environment. 
The system presented in this paper is a unique solution for the interaction of a swarm with dynamic obstacles.
The system is designed for a \ac{UAV} swarm and includes dynamic obstacle avoidance.
For mutual localization in the swarm, an onboard relative localization system with no communication requirements is used.
The system was deployed in the real world, as presented in the experimental section.

\section{Swarm system}
\label{sec_swarm}
\subsection{Localization}
The key for achieving compact flocking in nature - as well as for \acp{UAV} in real environments - is the knowledge of the relative states of the surrounding agents.
For collective evasion, the states of the interferers have to be retrieved as well.
The required relative state contains information about the relative position and velocity of team members within the range of localization sensors onboard a particular robot. 
The easiest approach for artificial systems is to implement communication between swarm agents and to broadcast the states of each agent in a global frame common to all swarm agents. 
In real-world environments, both the broadcasting in large groups and precise localization in a common frame are difficult to achieve with sufficient reliability \cite{LocalizationCommonFrame}.
Typically, the position estimation employs a global sensor such as \ac{GNSS}, which can easily lose signal in certain environments (e.g. forests, urban areas, indoors). 
Thus, its precision is insufficient for compact flocking~\cite{GPSaccuracy}.
The exact position is viable only under a precise localization system where all agents share the same coordinate frame, e.g. \ac{RTK}-\ac{GPS} \cite{rtk-gps}, which requires the presence of a pre-calibrated base station and a reliable \ac{GNSS} signal. 
Additionally, it is difficult to achieve reliability of communication (ad-hoc network) for the~variable size of the swarm and a sufficient transfer rate for up-to-date information about the surroundings in real-world conditions for large groups. 

Local sensing that is carried onboard robots for providing information about the proximity of other robots is a promising way of addressing these issues and is inspired by how organisms cooperate in nature.
The proposed swarming approach is designed to respect the limitations of onboard relative localization, such as the \ac{UVDAR} system designed by our group \cite{UVDAR, UVDAR1}, specifically for the intended compact swarming inspired by natural systems. 
The \ac{UVDAR} system consists of an onboard \ac{UVDAR} sensor (set of \ac{UV}-sensitive cameras) and \ac{UVDAR} blinking \ac{UV} \ac{LED} markers. 
The individual agents are recognized by unique blinking frequency of their markers and the relative positions of other \acp{UAV} are determined by bearing and distance estimation~\cite{UVDAR1}.
As the \ac{UVDAR} system provides relative positions, the velocity of neighboring agents is obtained by velocity estimation from time sequences of the observed positions.
The \ac{UVDAR} system exploits the properties of its \ac{UV} wavelength to significantly reduce dependence on lighting conditions when compared to classical vision-based alternatives.
Since no external infrastructure is required, the system can be used in real-world conditions including forests, urban areas, and indoor environments.
The interaction of humans with robots employing \ac{UVDAR} is safe due to the intensity and specific \ac{UV} wavelength it uses, as discussed in \cite{BioInspiredSwarm}.
In this paper, the \ac{UVDAR} system is used for \twoD{} localization.
However, an extension of localization to \threeD{} can be easily added, as presented in \cite{UVDAR, UVDAR1}.
The reason for flying the \acp{UAV} at the same altitude is to be able to study evasive behavior in a compact group of real \acp{UAV}, whose characteristics are best observed in \twoD{}.

\subsection{Swarm model}
\label{sec_swarm_model}
The basic swarm model proposed in this paper (to be used with systems such as \ac{UVDAR}) consists of three principles: keeping the swarm together (Flock centering or so called Cohesion), preventing collisions (mutual Collision avoidance via separation), and maintaining the direction and speed of the swarm (Velocity alignment).
This concept is a generalization of the well-known Boids model \cite{boids} that was chosen to demonstrate the general applicability of the proposed collective evasion approach.
These three basic rules are applied to neighbors of a swarm at distances smaller than $r_B$ from an agent. 
To describe the swarming model, let us define the state of each agent $j$ in terms of its position $\vec{p}_{j}$ and velocity $\vec{v}_{j}$ in the global frame as
\begin{align}
\vec{p}_{j} &= (x,y,z)^T,\\
\vec{v}_{j} &= (v_x,v_y,v_z)^T,
\end{align}
where both variables are vectors in $\mathbb{R}^3$.
The relative position~$\vec{p}_{j,i}  = \vec{p}_{i} - \vec{p}_{j}$ is a vector expressing the difference of the current position of a neighboring agent~$i$ and the current position of agent~$j$ obtained by the direct localization.

The first rule designed of the proposed model is applied to stabilize swarms of real \acp{UAV} with high dynamics and onboard sensors using uncertainty states where individual agents are attracted to each other within the distance $r_B$. 
This is enforced by the virtual Cohesion force, $\vec{F}_{C,j}$, the magnitude of which increases with the mutual distance of the agents from each other. 
The Cohesion force $\vec{F}_{C,j}$ is defined as
\begin{align}
\vec{F}_{C,j} = \dfrac{1}{n}\sum_{i=1}^n\epsilon_i \dfrac{\vec{p}_{j,i}}{\norm{\vec{p}_{j,i}}},
\end{align}
where $n$ is number of other agents that are located in the proximity of the agent $j$, and function $\epsilon_i$ is defined as
\begin{align}
\epsilon_i &= 
\begin{cases}
0, & p_{Ci} < d_{min},\\[0.2cm]
k_{1C} (p_{Ci} - d_{min})^2, & d_{min} < p_{Ci} < d_C,\\[0.2cm]
k_{2C}\log\left(k_{3C}(p_{Ci} - d_C) + 1\right) + \delta_C, & \text{otherwise,}\\
\end{cases}
\end{align}
where $p_{Ci} = \norm{\vec{p}_{j,i}} - l$ and $\delta_C = k_{1C} (d_C - d_{min})^2$.
The distance~$l$ is the safety distance around the agent $j$, $d_{min}$ is a distance in which the Cohesion force $\vec{F}_{C,j}$ starts to act, $d_C$ is a distance in which function $\epsilon_i$ is changed from a quadratic to logarithmic function, and $k_{1C}$ with physical units m$^{-2}$ and $k_{2C}$ are scaling factors in function $\epsilon_i$. Term $k_{3C} = 1$~m$^{-1}$ is applied to scale the result of subtraction $(p_{Ci} - d_{C})$.
In comparison with most other works focused on swarming, we consider the physical units in the swarm model allowing us to link it with \ac{UAV} motion constraints and properties of onboard localization constraints.

The Cohesion force $\vec{F}_{C,j}$ is designed to smooth convergence of a quadratic function in the neighborhood of the local minimum. 
If the relative distance between two agents grows to infinity, the quadratic function is replaced by the logarithmic function to slow the growth.  
In case the relative distance between two agents is approaching zero, the Cohesion force also becomes zero.

The second rule prevents collisions among the agents and protects them from getting too close to each other. 
If the mutual distance between two agents is below a threshold, the virtual Separation force $\vec{F}_{S,j}$ becomes active and its magnitude increases with the decreasing distance between agents. 
The direction of this force is opposite to the direction of the Cohesion force $\vec{F}_{C,j}$ associated with the same neighbor. 
To determine its magnitude for one agent $j$, it is necessary to know the relative positions $\vec{p}_{j,i}$ of the surrounding $n$ swarm agents.
In the Separation force
\begin{align}
\vec{F}_{S,j} = -\dfrac{1}{n}\sum_{i=1}^n\kappa_i \dfrac{\vec{p}_{j,i}}{\norm{\vec{p}_{j,i}}},
\label{SeparationForce}
\end{align}
proposed for real \ac{UAV} swarms, the function $\kappa_i$ is defined as
\begin{align}
\kappa_i &= 
\begin{cases}
0, & d_{max} < p_{Si},\\[0.2cm]
k_{1S} (p_{Si} - d_{max})^2, & d_S < p_{Si} < d_{max},\\[0.2cm]
k_{2S} \left( \dfrac{\sqrt{p_{Si}}}{p_{Si}} - \dfrac{\sqrt{d_{max}}}{d_{max}} \right) + \delta_S, & \text{otherwise},\\
\end{cases}
\label{eq:sepForceKappai}
\end{align}
where
\begin{align}
p_{Si} &= \begin{cases}
\norm{\vec{p}_{j,i}} - l, & \norm{\vec{p}_{j,i}} > l+l_{min},\\[0.2cm]
l_{min}, & \text{otherwise,}\\
\end{cases}
\end{align}
\begin{align}
\delta_S &= k_{1S}(d_S - d_{max})^2 - k_{2S}\left( \dfrac{\sqrt{d_S}}{d_S} - \dfrac{\sqrt{d_{max}}}{d_{max}} \right).
\end{align}
Parameter $l_{min}$ is the minimal value of function $p_{Si}$, $d_{max}$ is the maximal distance in which the Separation force $\vec{F}_{S,j}$ is active, $d_S$ is a distance in which function $\kappa_i$ is changed to a quadratic function, and the scaling factors are $k_{1S}$ with physical unit m$^{-2}$ and $k_{2S}$ with physical unit m$^{\frac{1}{2}}$.

The Separation force $\vec{F}_{S,j}$ was designed based on the requirement of smooth convergence in the neighborhood of the local minimum using the quadratic function. 
In the event that the relative distance between two agents decreases to zero, the quadratic function is substituted with a steeper function (\ref{eq:sepForceKappai}) to increase the growth rate.
If the relative distance between two agents grows to infinity, the quadratic function is set to zero to stop agents from moving apart.

The last rule applies the virtual Alignment force $\vec{F}_{A,j}$ that takes into account the velocity of the surrounding agents of the swarm.
In the original Boids model, the force affects the orientation of the agents, but in the case of real \acp{UAV} with omnidirectional movement capabilities, we apply it to the movement direction to achieve the required stable swarming behavior.
The Alignment force $\vec{F}_{A,j}$ unifies the direction of movement of all individuals in the swarm and also decreases the likelihood of collisions, since it forces the direction of the movement of all the agents in the swarm to become parallel. This is also important for achieving stability of real robots. 
We propose the Alignment force $\vec{F}_{A,j}$ as 
\begin{align}
\vec{F}_{A,j} = \dfrac{1}{n}\sum_{i=1}^n k_A \vec{v}_i,
\label{AlignmentForce}
\end{align}
where the parameter $k_A$ with the physical unit s$\cdot$m$^{-1}$ is a scaling factor.

The Total force $\vec{F}_{T,j}$ acting on an agent $j$ is given by combining all of the presented forces of Cohesion, Separation, and Alignment as
\begin{equation}
\vec{F}_{T,j} = \vec{F}_{C,j} + \vec{F}_{S,j} + \vec{F}_{A,j}.
\label{eq:totalForce}
\end{equation}
Note, that all presented forces are normalized as dimensionless quantities used to determine the desired acceleration of each agent at a given time.
The acceleration is proportional to the magnitude of these forces (as would be the case with classical physical forces) with the factor of a unit virtual mass $m = 1$. 
For the physical \ac{UAV} swarm agents control, we propose using the acceleration $\vec{a}_j=\vec{F}_{T,j}/m$ to determine the newly desired position $\vec{p}_{j}^{~k+1}$ with the current position $\vec{p}_{j}^{~k}$.
The desired position $\vec{p}_{j}^{~k+1}$ of the agent $j$ is determined considering the linear motion of the agent as
\begin{align}
\vec{p}_{j}^{~k+1} = \vec{p}_{j}^{~k} + \vec{v}_j(t_{c+1} - t_{c}) + \dfrac{1}{2}k_{pa}\vec{a}_j(t_{c+1} - t_{c})^2,
\label{desiredPosition}
\end{align}
where $t_{c}$ is the time of the last swarm control step~$c$ and $t_{c+1}$ is the time of the new control step $c+1$.
Terms $k_{pv} = 1$ m$\cdot$s$^{-1}$ and $k_{pa} = k_{pv} k_{va} = 1$ m$\cdot$s$^{-2}$ are applied to achieve physical meaning of the obtained acceleration to be placed directly into \ac{UAV} control rules. 
This is a novel approach facilitating understanding of ongoing swarming behavior in real-world conditions.
Finally, the position $\vec{p}_{j}^{~k+1}$ is passed to the \ac{MPC} controller that has been proposed in \cite{MPC} to allow integration of swarming rules into a stable swarm relying exclusively on onboard sensing.

\section{Fast collective evasion}
The proposed fast collective evasion for avoiding moving objects (interferers) approaching the swarm consists of three states (evasive modes) -- Normal mode, Active mode and Passive mode. 
In the Normal mode, the agent does not detect an approaching interferer and other agents do not report the presence of an interferer. 
In the Active mode, an agent detects an interferer and starts to escape.
When an agent does not see an interferer, but the other agents report the presence of an interferer, the state is switched to the Passive mode.

\subsection{Normal mode}
\begin{figure}[!htb]
\centering
\includegraphics[width=0.55\columnwidth]{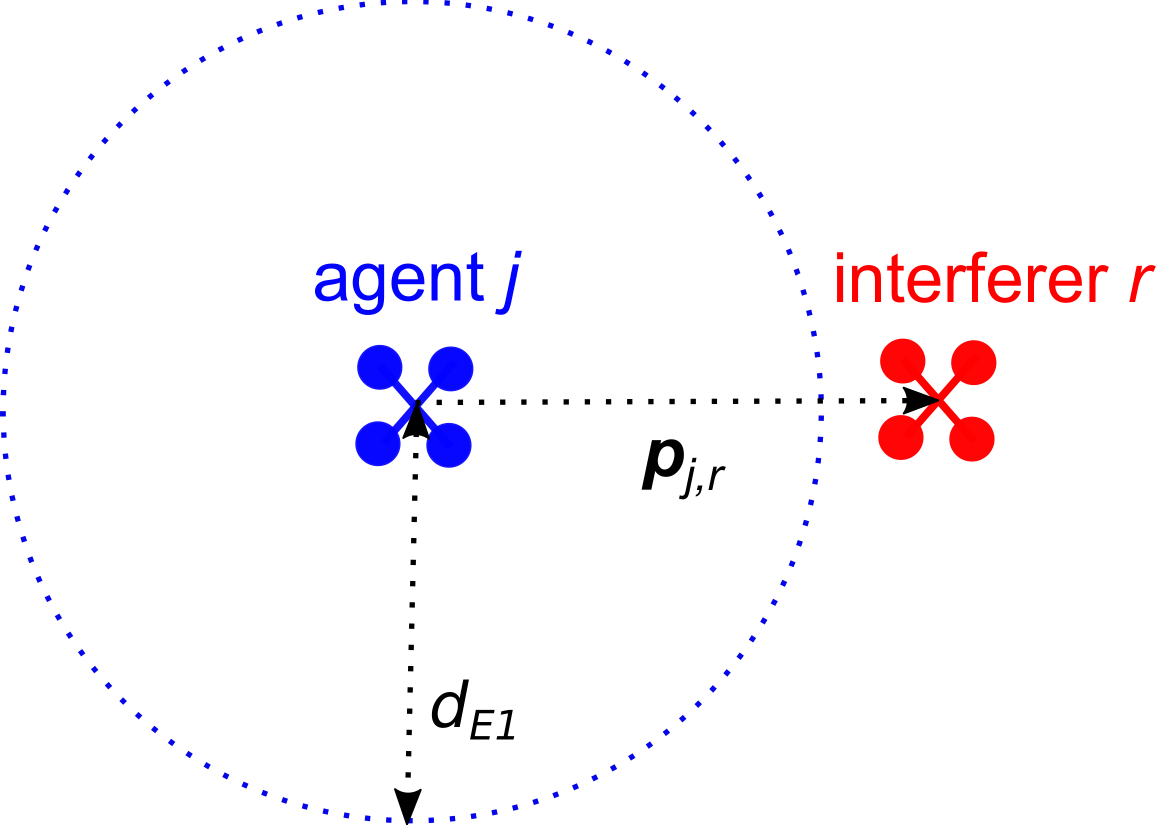}
\caption{Visualization of the Normal mode - the agent $j$ in Normal mode is denoted in blue and the interferer is red.}
\label{fig:schemeNormalMode}
\end{figure}

In the Normal mode, the sensors that the agent $j$ uses to observe its surroundings do not reveal any object considered an interferer~$r$ or the distance $\norm{\vec{p}_{j,r}}$ to a potential interferer is greater than a threshold $d_{E1}$ (Figure \ref{fig:schemeNormalMode}). 
Additionally, the other swarm agents do not report the presence of an interferer.
The movement of an agent in Normal mode is controlled only by the basic swarm model presented in section \ref{sec_swarm_model}. 
The newly desired position is calculated by equation (\ref{desiredPosition}).

\subsection{Active mode}
\begin{figure}[!htb]
\centering
\includegraphics[width=0.4\columnwidth]{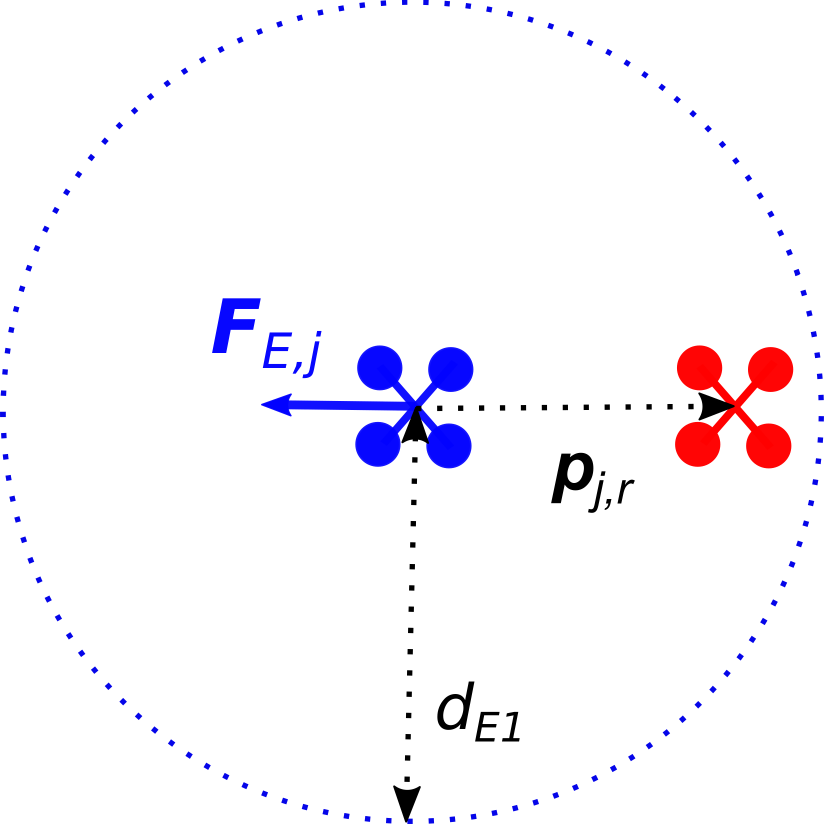}
\caption{Visualization of the Active mode - the agent $j$ in Active mode is denoted in blue and the interferer is red.}
\label{fig:schemeActiveMode}
\end{figure}

When the agent $j$ detects an interferer at a distance smaller than $d_{E1}$, its state is switched to the Active mode in order to move the agent $j$ away from the interferer to safety.
This is implemented by adding a new virtual force called the Escape force $\vec{F}_{E,j}$, which is based on \cite{MinWang}.
The magnitude of $\vec{F}_{E,j}$ increases with decreasing distance between the agent and the interferer. 
The direction of $\vec{F}_{E,j}$ is away from the interferer (Figure \ref{fig:schemeActiveMode}).

The Escape force $\vec{F}_{E,j}$ for $m$ interferers is obtained as
\begin{equation}
\vec{F}_{E,j} = -\dfrac{1}{m}\sum_{r=1}^m\sigma_r~\dfrac{\vec{p}_{j,r}}{\norm{\vec{p}_{j,r}}},
\end{equation}
where $\vec{p}_{j,r}$ denotes the relative position between the agent $j$ and the interferer $r$. 
The function $\sigma_r$ is defined as
\begin{align}
\sigma_r &= 
\begin{cases}
k_E\left(\dfrac{\sqrt{p_{Er}}}{p_{Er}} - \dfrac{\sqrt{d_{E2}}}{d_{E2}} \right), & p_{Er} < d_{E2},\\[0.2cm]
0, & \text{otherwise,}
\end{cases}
\label{sigma_r}
\end{align}
where
\begin{align}
p_{Er} &= \begin{cases}
\norm{\vec{p}_{j,r}} - l, & \norm{\vec{p}_{j,r}} > l+l_{min},\\[0.2cm]
l_{min}, & \text{otherwise.}
\end{cases}
\end{align}
The parameter $k_E$ with physical unit m$^{\frac{1}{2}}$ is the scaling factor and $d_{E2}$ is a distance in which the Escape force $\vec{F}_{E,j}$ stops contributing to the decentralized control law.
The function $\sigma_r$ (\ref{sigma_r}) is designed to increase the magnitude of the Escape force $\vec{F}_{E,j}$ in the event of a decreasing distance between the agent $j$ and interferers.

The Total force $\vec{F}_{T,j}$ is then obtained as
\begin{equation}
\vec{F}_{T,j} = \vec{F}_{C,j} + \vec{F}_{S,j} + \vec{F}_{A,j} + \vec{F}_{E,j}.
\end{equation}
Again, note that all presented forces are dimensionless quantities used to compute the virtual acceleration $\vec{a}_j$, as shown in section \ref{sec_swarm_model}, assuming virtual mass $m=1$. 
The virtual acceleration $\vec{a}_j$ is then used to determine the new desired position according to equation (\ref{desiredPosition}). 

\subsection{Passive mode}
\begin{figure}[!htb]
\centering
\includegraphics[width=0.4\columnwidth]{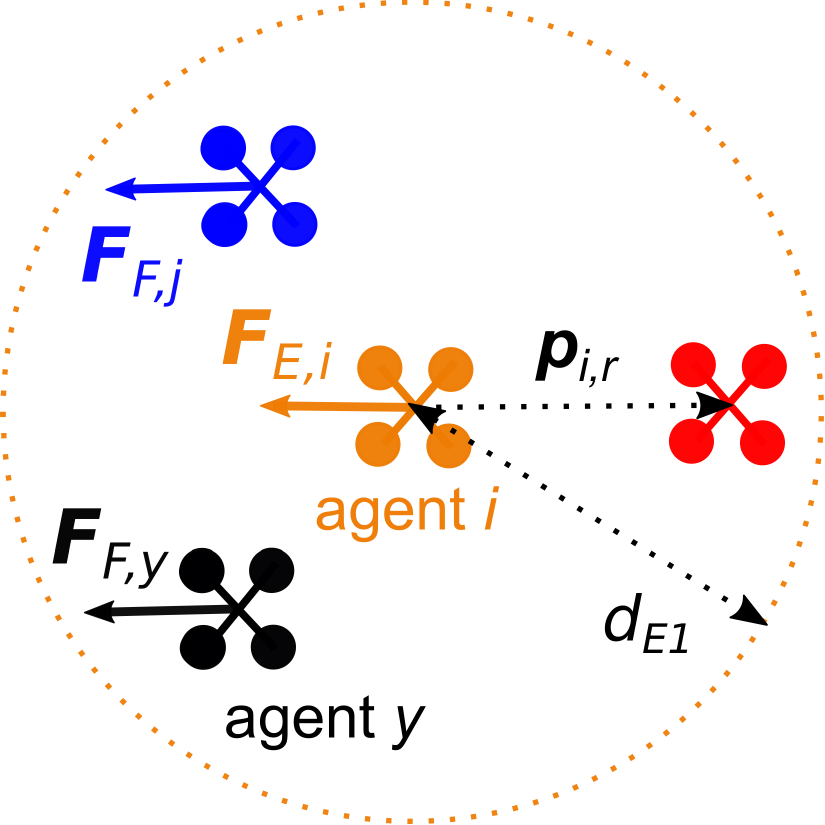}
\caption{Visualization of the Passive mode - the orange agent $i$ is in Active mode, the blue agent $j$ and black agent $y$ are in Passive mode, and the interferer is shown in red.}
\label{fig:schemePassiveMode}
\end{figure}

Even if an agent $j$ does not detect an interferer, it does not mean that the interferer may not be located nearby. 
If another agent of the swarm $i$ can detect the interferer, it starts to escape using the Escape force $\vec{F}_{E,i}$ and it reports the presence of an interferer to other swarm agents simultaneously.
If the agent $j$ receives such a report about the presence of an interferer, its escape state is changed to the Passive mode (Figure \ref{fig:schemePassiveMode}).
In nature, such shock propagation is triggered by the detection of sudden and highly dynamic motion of neighbors.
In a \ac{UAV} swarm, the same principle of direct observation of the neighbors' states can be applied. 
However, due to high uncertainty of velocity and acceleration measurements provided by current mutual localization sensors, we recommend using an additional signal, such as changing the \ac{UVDAR} \cite{UVDAR, UVDAR1} frequency to trigger the transition between the evasive modes.

While the Normal mode provides more freedom to swarm particles for tackling a task given by a particular application, the swarm interaction is increased in the Passive mode.
The agents are forced to react more dynamically to the movement of others with less priority given to the intended task.
The priority in the Passive mode is given to collective evasion from an interferer in a similar way as is observed in natural swarm systems \cite{predatorSwarmInteraction}.
To achieve a faster response and closer interaction, a new virtual force called the Following force $\vec{F}_{F,j}$ will be introduced. 
The purpose of the Following force $\vec{F}_{F,j}$ is to unify the direction of movement of swarm agents with a quicker response than the separate Alignment force $\vec{F}_{A,j}$ (\ref{AlignmentForce}) provides.
Moreover, the parameters of the Cohesion force $\vec{F}_{C,j}$ and of the Separation force $\vec{F}_{S,j}$ will be modified to decrease the reaction time.

The Following force $\vec{F}_{F,j}$ is proposed based on the velocity $\vec{v}_i$ of other~$n$ swarm agents as
\begin{equation}
\vec{F}_{F,j} = \dfrac{1}{n}\sum_{i=1}^n\zeta_i \dfrac{\vec{v}_i}{\norm{\vec{v}_i}}.
\end{equation}
The function $\zeta_i$ is defined as
\begin{equation}
\zeta_i = k_F\log(k_{V}\norm{\vec{v_i}} + d_F),
\end{equation}
where parameter $k_{V}$ with physical unit s$\cdot$m$^{-1}$ and $k_{F}$ and $d_F$ are scaling factors of function $\zeta_i$.

The Total force acting on an agent in the Passive mode is
\begin{equation}
\vec{F}_{T,j} = \vec{F}_{C,j} + \vec{F}_{S,j} + \vec{F}_{A,j} + \vec{F}_{F,j}.
\end{equation}
The newly desired position of an agent is determined as it is in Normal mode, based on equation (\ref{desiredPosition}).

\subsection{Transition between modes using low-bandwidth communication}
\label{transitionWithCommunication}
Let us discus the overall evasive behavior composed from the modes described above.
In a standard situation, the agent~$j$ is controlled under the Normal mode.
If agent~$j$ detects an interferer within the distance of $d_{E1}$, the mode of agent~$j$ is changed to the Active mode.
The agent~$j$ then begins the evasive action. 
The agents in local proximity are informed about this transition by means of \ac{UVDAR} with changes in the frequency of the \ac{UV} \acp{LED}. 
If a direct local communication is allowed, a message containing the name of the agent~$j$, timestamp, and information about the detected interferer is sent to all surrounding agents within the communication range.

In case an agent~$i$ has not detected an interferer within the distance of $d_{E1}$ and has not observed or received any information about an agent being in Active mode yet, the agent~$i$ stays in the Normal mode.
If any received message reports that another agent~$j$ is escaping from an interferer, the agent $i$ changes its mode to the Passive mode. 
The agent~$i$ stores the name of the agent~$j$ from the message and the message is forwarded to its neighbors.
When the agent~$i$ receives a message with the already stored name, the message is not forwarded.

If the agent~$j$ stops detecting the interferer or the distance to the interferer is greater than $d_{E2}$, the agent~$j$ sends a message that the interferer is not present to all surrounding agents within communication range.
The mode of the agent~$j$ is then switched back to the Normal mode only if an array of stored names of the agent~$j$ is empty.
Otherwise, the mode of the agent~$j$ is changed to the Passive mode.
Additionally, consider the agent~$j$ received a message about the detected interferer with its own name. 
However, the agent~$j$ is in the Normal mode.
The message that the interferer is not present is then sent to all surrounding agents within the communication range once more. 
This mechanism is designed to increase system reliability, which is highly important in the case of implicit communication in large swarms.

In the event that the agent~$i$ received a message that the interferer is not present, the corresponding stored name is erased from its list. 
If the agent~$i$ has not stored any other name, the mode of the agent~$i$ is changed to the Normal mode.
When conflicting messages are received, i.e. the first message reports that the agent~$j$ is escaping and the second message reports that no interferer is detected by the agent~$j$, the message with a greater timestamp is taken into account and the other message is discarded.
Again, such an approach is crucial for large swarms in real-world conditions where the properties of communication mesh are difficult to estimate for strictly decentralized simple swarming agents.

\begin{figure}[!htb]
\centering
\includegraphics[width=\columnwidth]{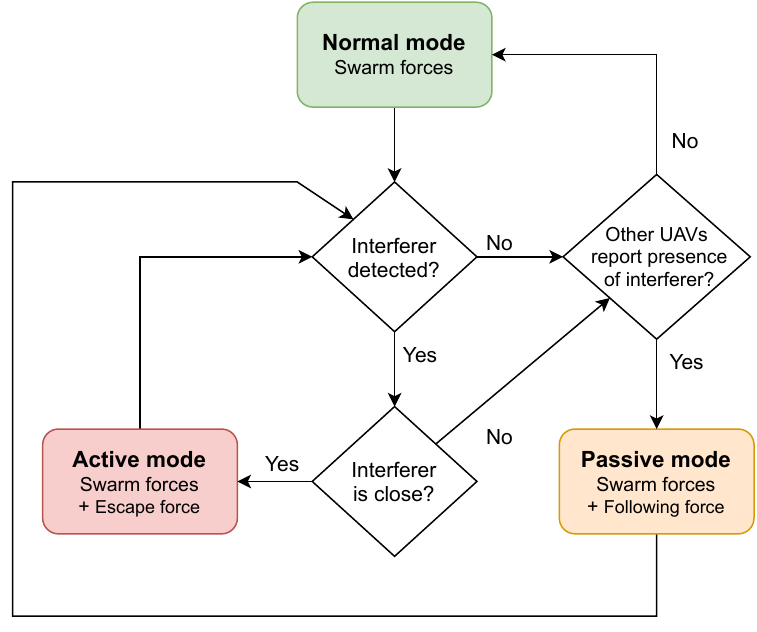}
\caption{Diagram of the collective evasive behavior.}
\label{fig:diagramEscapeBehaviour}
\end{figure}

The diagram in Figure \ref{fig:diagramEscapeBehaviour} shows the structure of the collective evasion behavior presented in this paper.
The term Swarm forces in the diagram denotes the Total force $\vec{F}_{T,j}$ from the equation (\ref{eq:totalForce}) retrieved by the swarm model from section \ref{sec_swarm_model}.
\begin{figure*}
 	\centering
	\subfigure{
		\includegraphics[width=0.47\columnwidth]{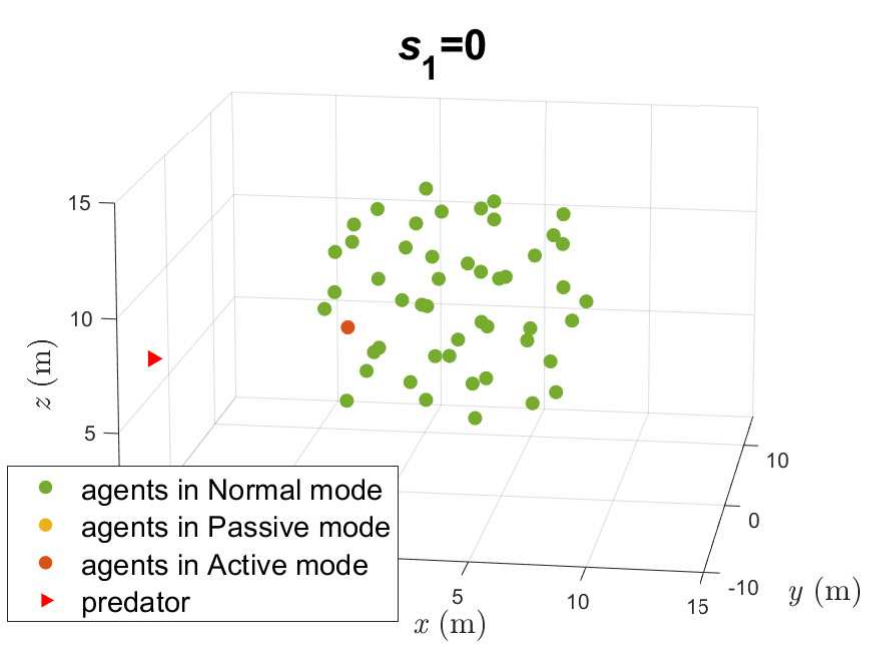}
	}
	\subfigure{
		\includegraphics[width=0.47\columnwidth]{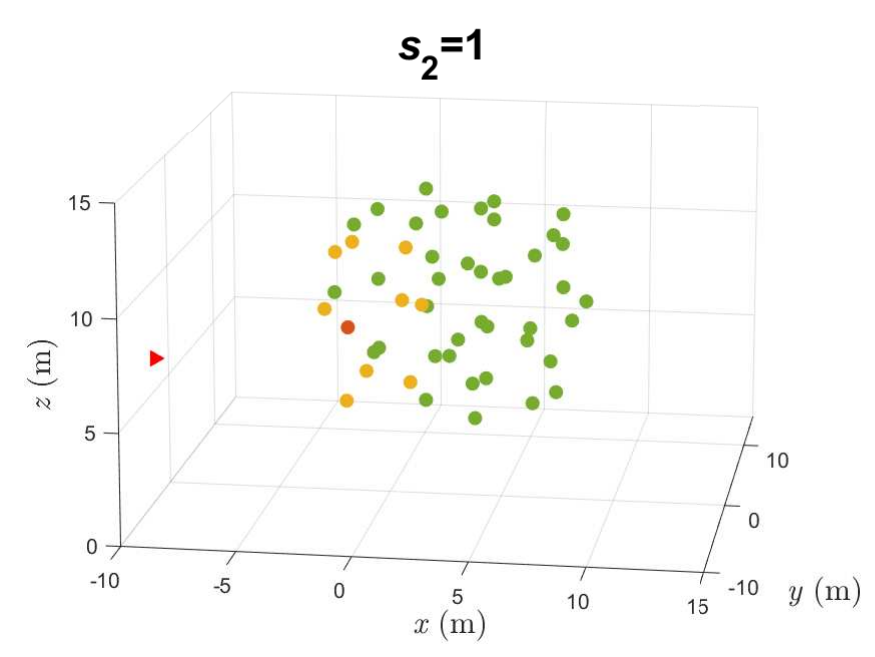}
	}
	\subfigure{
		\includegraphics[width=0.47\columnwidth]{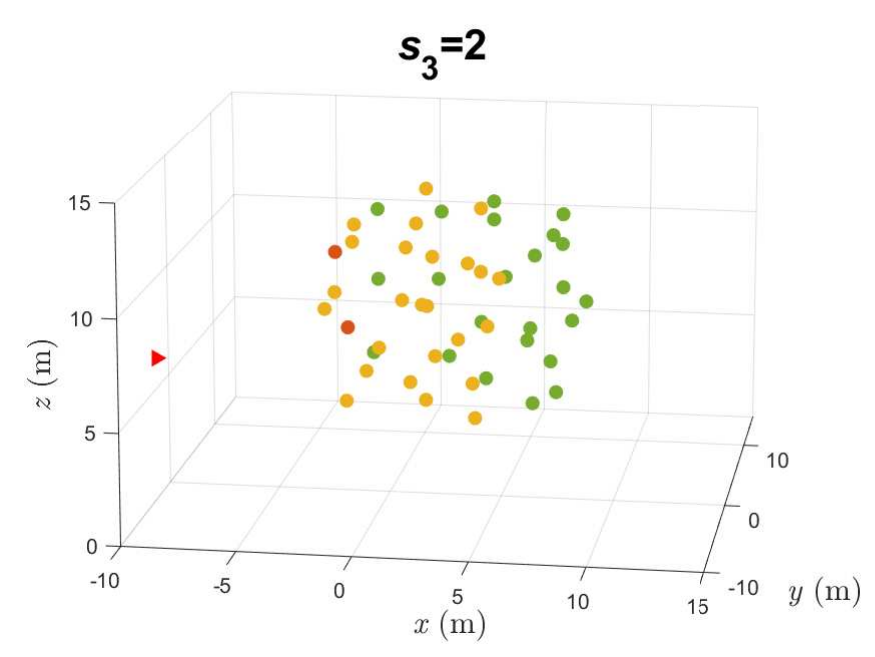}
	}
	\subfigure{
		\includegraphics[width=0.47\columnwidth]{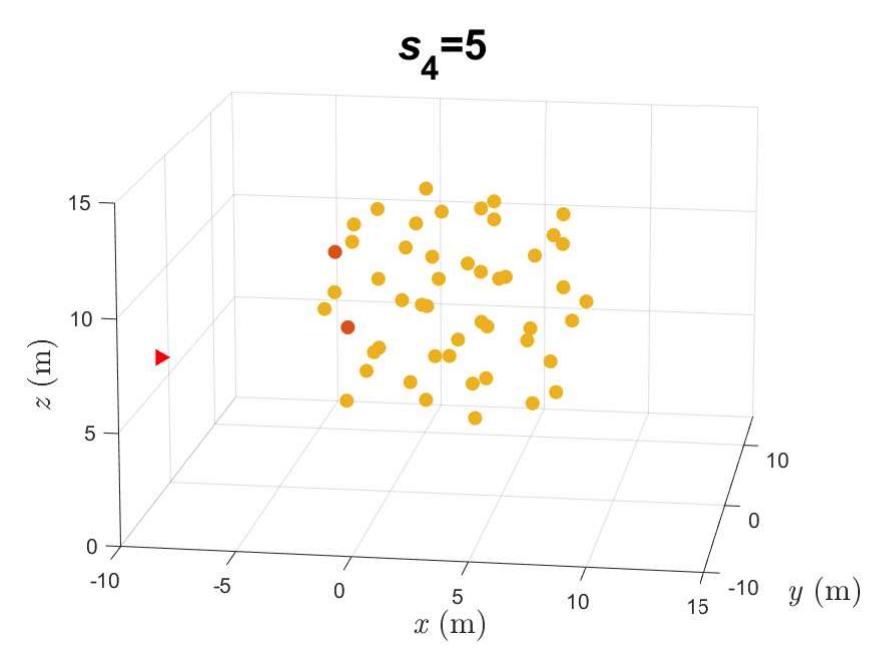}
	}
	\caption{Shock propagation in a swarm of 50 agents when an interferer is detected by a swarm agent.}
	\label{shockProp}
\end{figure*}
\begin{figure*}
 	\centering
	\subfigure{
		\includegraphics[width=0.47\columnwidth]{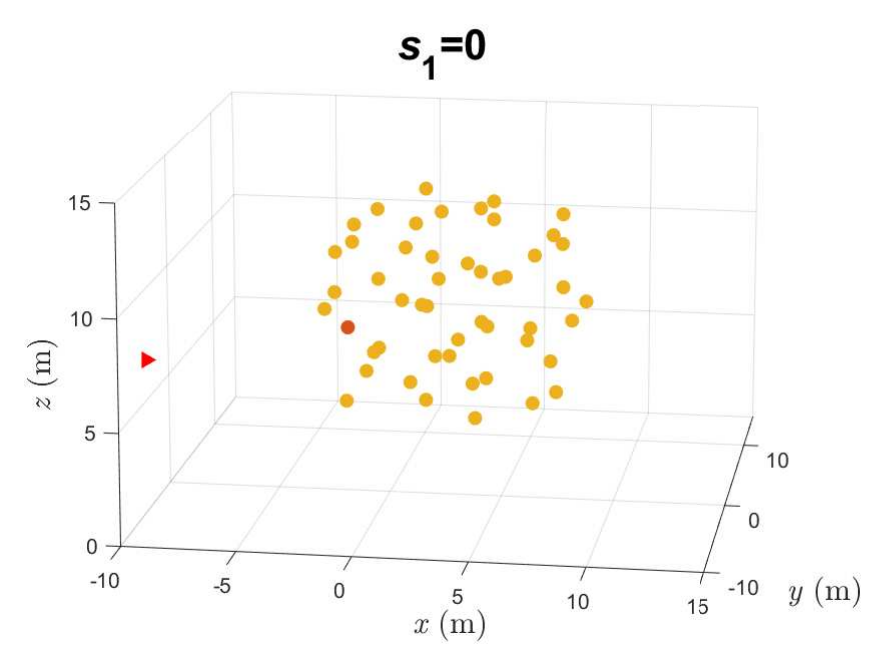}
	}
	\subfigure{
		\includegraphics[width=0.47\columnwidth]{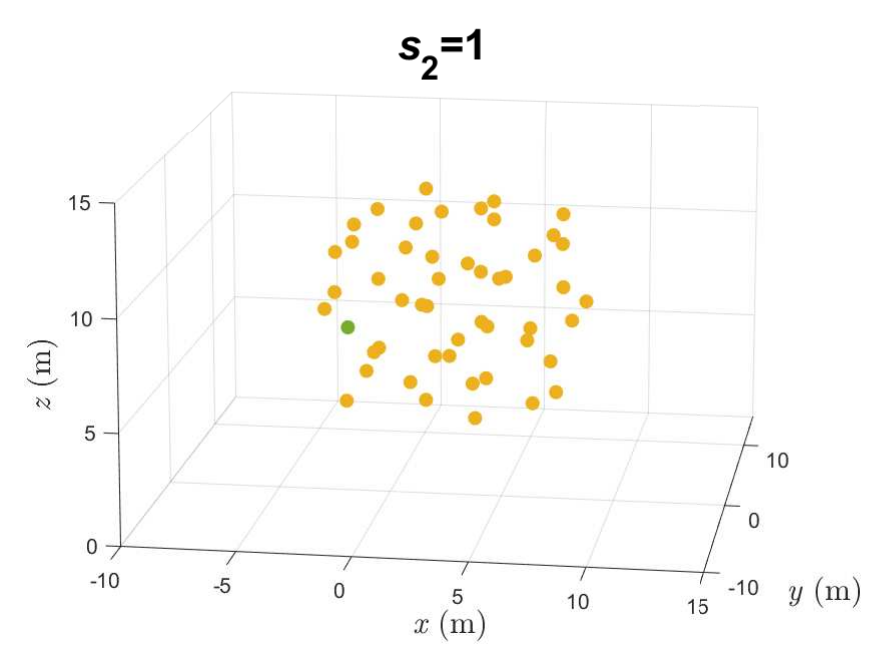}
	}
	\subfigure{
		\includegraphics[width=0.47\columnwidth]{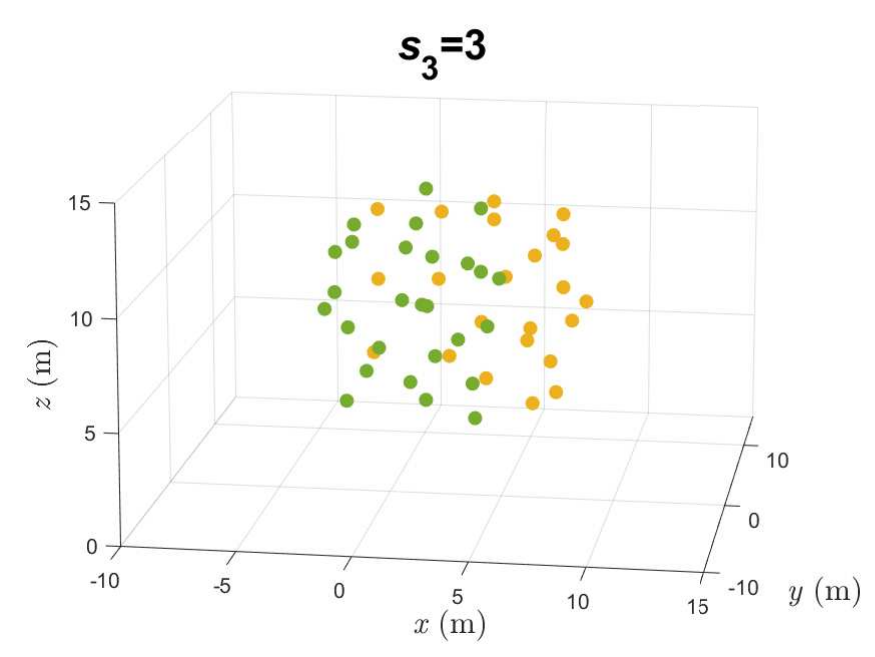}
	}
	\subfigure{
		\includegraphics[width=0.47\columnwidth]{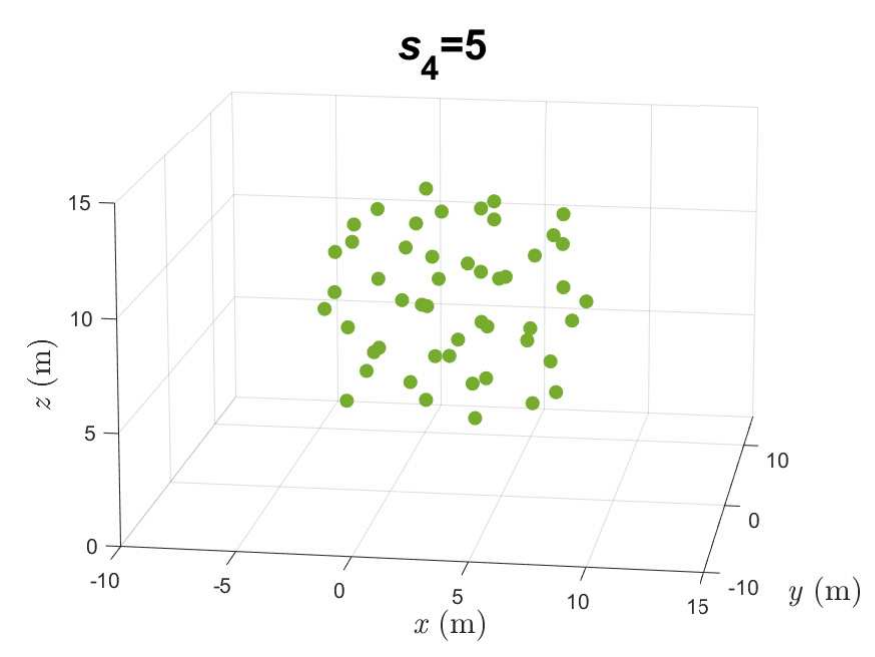}
	}
	\caption{Shock propagation in a swarm of 50 agents when a detected interferer achieved a safe distance from the swarm.}
	\label{shockBackProp}
\end{figure*}

\section{Experimental evaluation}
\subsection{Matlab simulations}
The system described in this paper was initially evaluated in a simplistic kinematic simulation to demonstrate the proposed transition between evasive modes in large \ac{UAV} groups. 
As shown in Figures \ref{shockProp} and \ref{shockBackProp} in open space, the agents create a shape similar to a sphere using swarm rules presented in section~\ref{sec_swarm_model}.
Once the swarm reaches a stable equilibrium, the swarm agents are frozen in their current static positions in order to focus only on shock propagation.
The mean distance between swarm agents is \SI{2.89}{\meter}, the minimal distance is \SI{2.45}{\meter}, and the maximal distance is \SI{3.46}{\meter}.
In the simulation, the transfer of one message via a network from an agent to its neighbors located at a distance equal to or less than 5 m takes one simulation step $s$.

The shock propagation - the transfer of information about a detected interferer - in a swarm of 50 agents is shown in Figure~\ref{shockProp}. 
The figure shows how the information about the detected interferer is propagated in the swarm when an interferer is detected by a swarm agent.
All of the swarm agents become aware of the presence of the detected interferer after five simulation steps.
Figure \ref{shockBackProp} shows the return of the swarm agents to Normal mode once the interferer is at a safe distance from the swarm.
All agents within the swarm return to this mode after five simulation steps, which is similar to the time required for spreading information about a newly detected interferer.
The same progress of shock propagation was also observed in the case of a moving group using the proposed approach and is shown in the following experiments.

\subsection{Simulation -- without evasive behavior}
\label{sec:GazeboOnlyBoids}
\begin{figure}[!b]
\centering
\includegraphics[width=0.95\columnwidth]{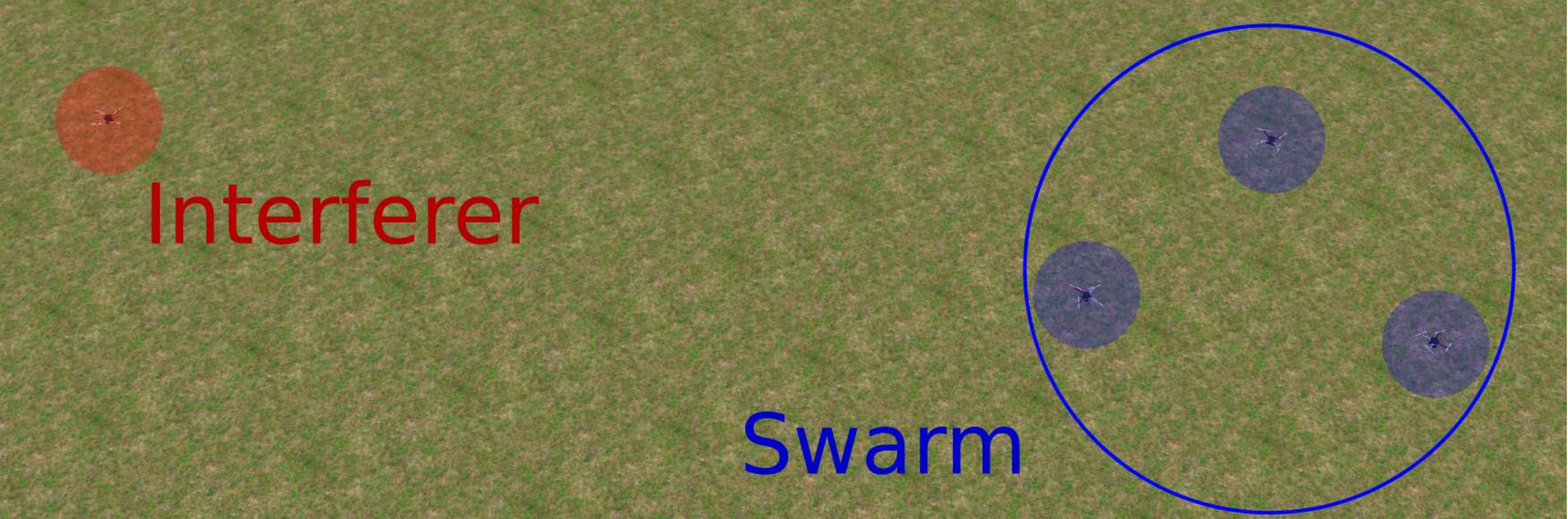}
\caption{Simulation with four \acp{UAV} in the Gazebo simulator.}
\label{fig:4UAVsGazebo}
\end{figure}

To imitate more realistic conditions, the system was subsequently tested in the Gazebo simulator. 
Since the requirements on computational resources are significantly higher in Gazebo with the simulated \ac{UVDAR} system \cite{UVDAR, UVDAR1}, only a swarm with a limited number of agents could be tested, i.e. three \acp{UAV} in a swarm and one \ac{UAV} as the interferer.
An example snapshot from one of these simulations is shown in Figure \ref{fig:4UAVsGazebo}.

In the first simulation in~Gazebo, the proposed evasive behavior is compared with the basic swarm model (see video in~\footnote{\url{http://mrs.felk.cvut.cz/papers/uav-swarm-fast-collective-evasion}\label{footnote-videos}}).
If just the swarm model is used, an interferer can be detected as only a static obstacle to be considered in a similar way as to other swarm members in the Separation force of equation~(\ref{SeparationForce}).
Only the detection range can be enlarged to respect the maximum detection range of the onboard vision system.

\begin{figure}[!htb]
\centering
\includegraphics[width=0.8\columnwidth]{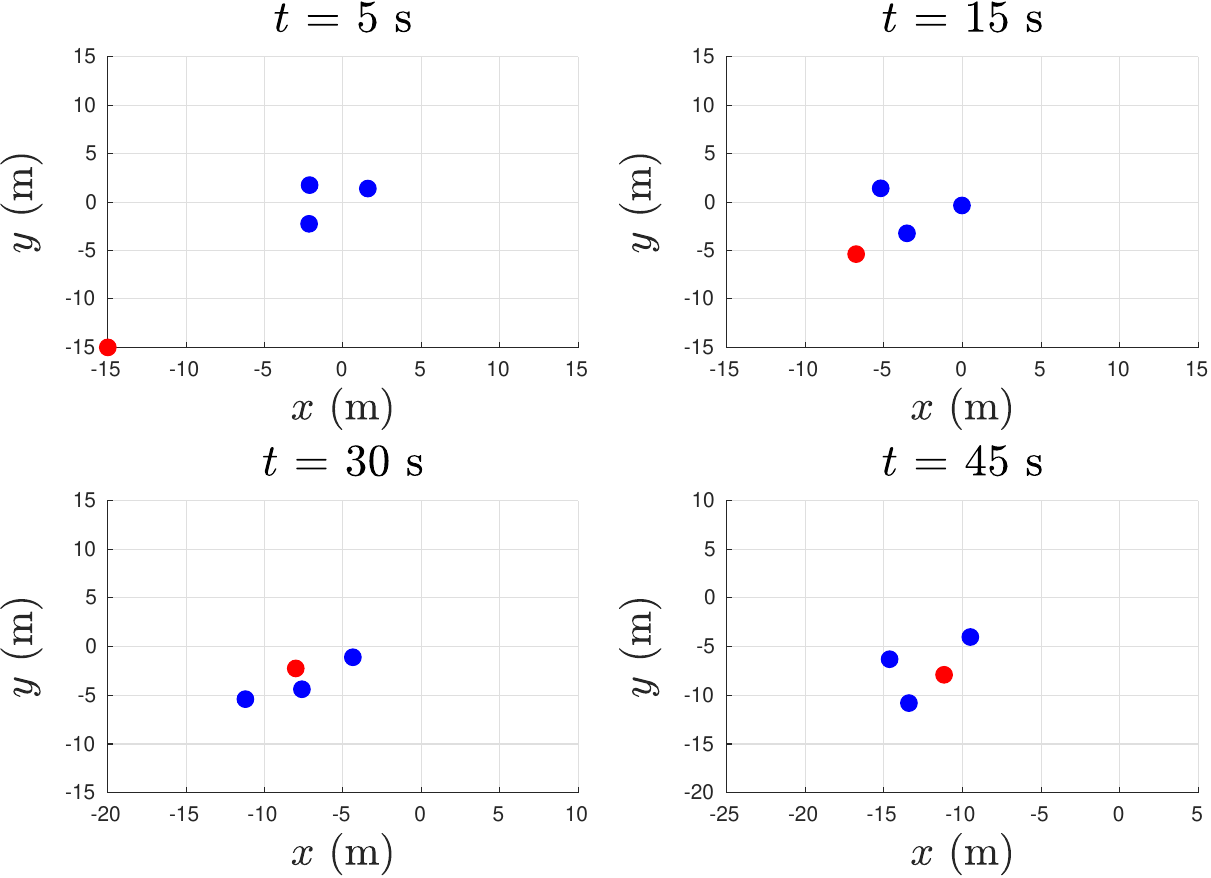}
\caption{Movement of swarm \acp{UAV} (blue) and an interferer (red) in the simulation without using evasive behavior, as in the experiment introduced in Figure \ref{fig:4UAVsGazebo}.}
\label{fig:withoutEvasiveBehaviorPositions}
\end{figure}

\begin{figure}[!htb]
\centering
\includegraphics[width=0.85\columnwidth]{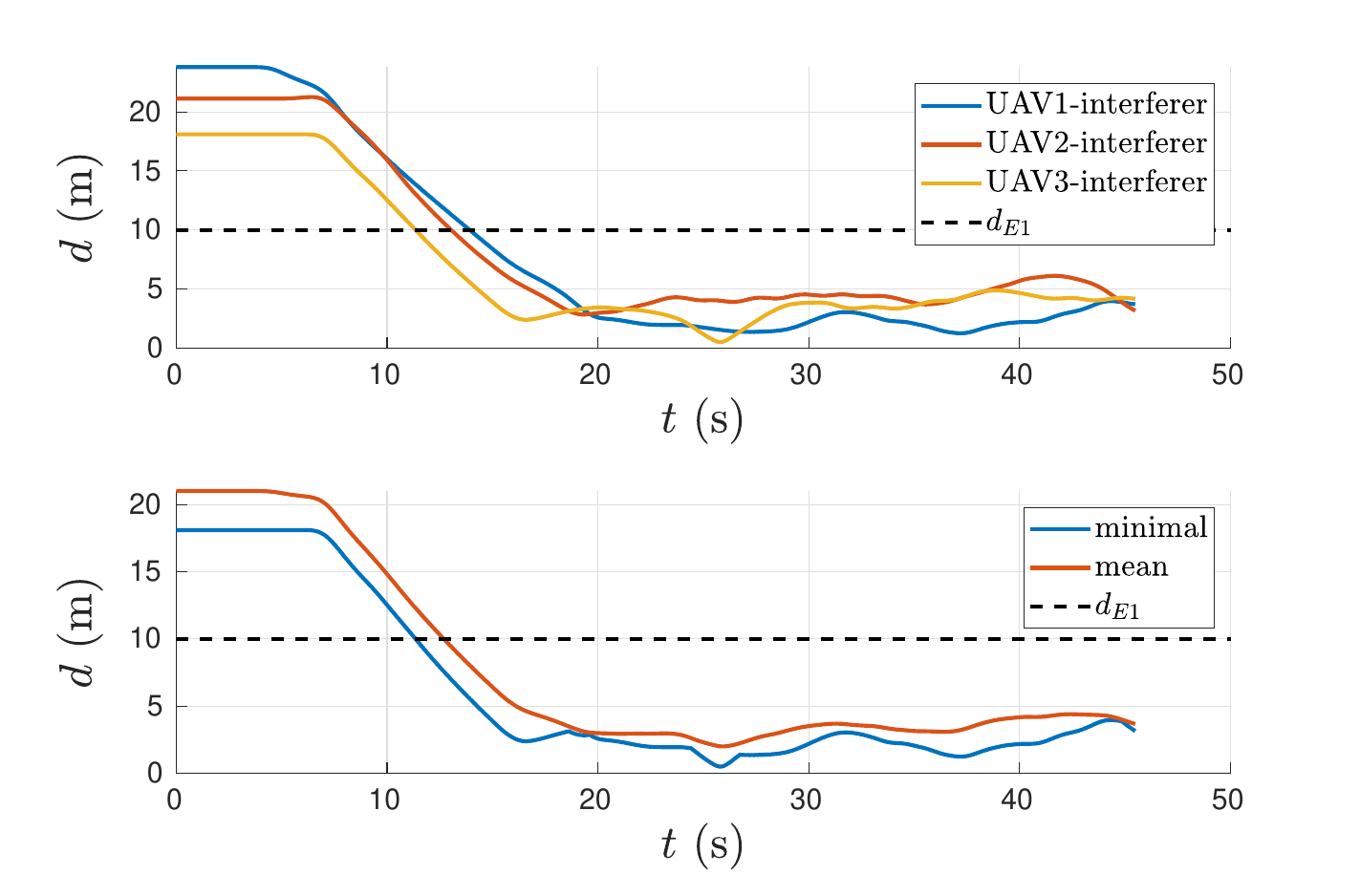}
\caption{The upper figure is a graph of distances between swarm \acp{UAV} and the interferer if all swarm \acp{UAV} do not use evasive behavior, as in the experiment introduced in Figure \ref{fig:4UAVsGazebo}. 
The lower figure shows the minimal and mean distance between swarm \acp{UAV} and the interferer.}
\label{fig:withoutEvasiveBehaviorDistPred}
\end{figure}

\begin{figure}[!htb]
\centering
\includegraphics[width=0.715\columnwidth]{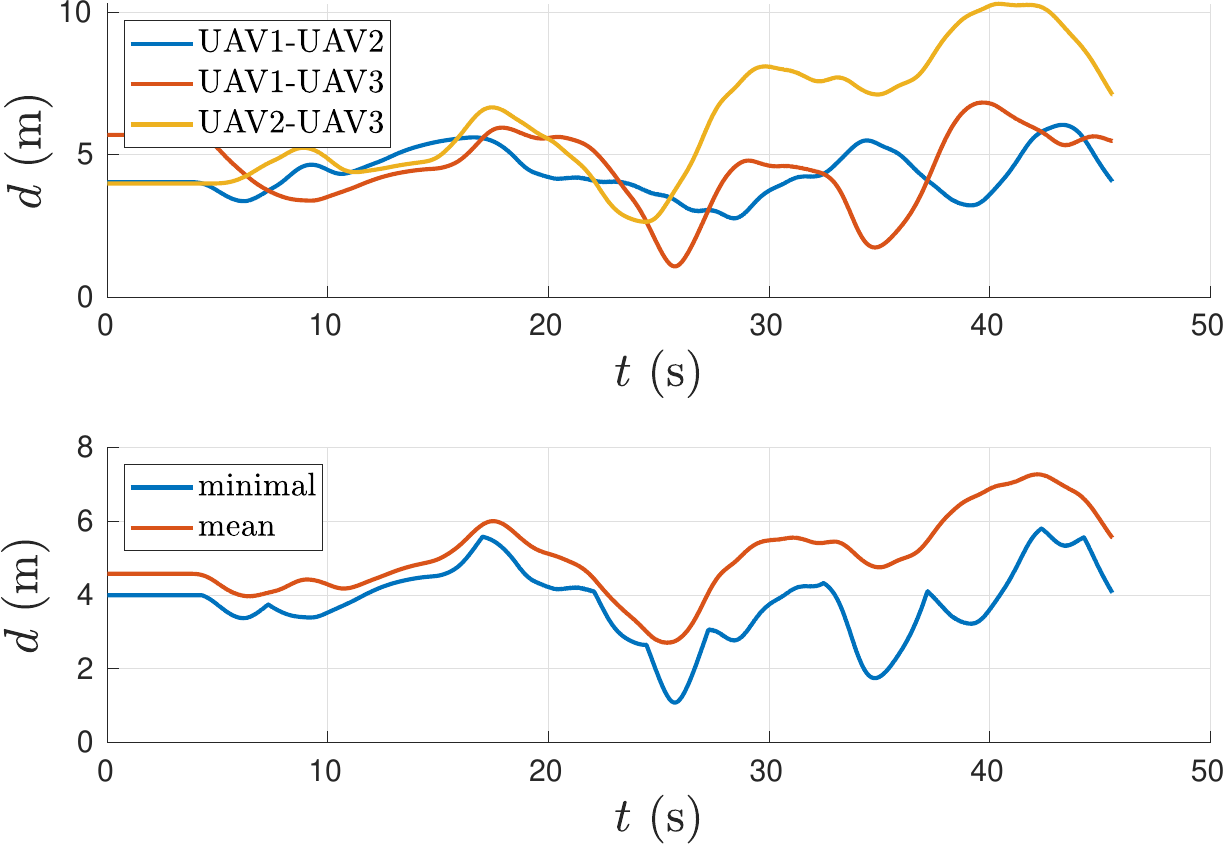}
\caption{The absolute, minimal, and mean distances between swarm \acp{UAV} if all swarm \acp{UAV} do not use evasive behavior, as in the experiment introduced in Figure \ref{fig:4UAVsGazebo}.}
\label{fig:withoutEvasiveBehaviorDistUAVs}
\end{figure}

The positions of the swarm \acp{UAV} in blue and the interfer in red at $x,~y$ coordinates at a time~$t$ are shown in Figure~\ref{fig:withoutEvasiveBehaviorPositions}.
The sub-figures present an interesting behavior. 
When the swarm \acp{UAV} do not use the fast collective evasion forces, they flock around the interferer.
The behavior of the interferer and the swarm \acp{UAV} becomes similar to a situation where the interferer would act as another swarm agent. 
The~most straightforward improvement possible is to enlarge the separation force, but this does not present a sufficient effect, as shown in \cite{BioInspiredSwarm}.
The Figure \ref{fig:withoutEvasiveBehaviorDistPred} displays distances between the swarm \acp{UAV} and the interferer. 
This figure demonstrates that the distance between the swarm and the interferer first diminishes and then stabilizes at approx. \SI{3}{\meter}.
Lastly, Figure \ref{fig:withoutEvasiveBehaviorDistUAVs} shows the distances between the swarm \acp{UAV}.
The minimum of these distances drops to \SI{1}{\meter}, which can be a hazardously close distance between individual swarm \acp{UAV}.


\subsection{Simulation -- without evasive behavior -- enlarged detection range}
\label{sec:GazeboOnlyBoidsEnlargedDetectionRange}
\begin{figure}[!htb]
\centering
\includegraphics[width=0.9\columnwidth]{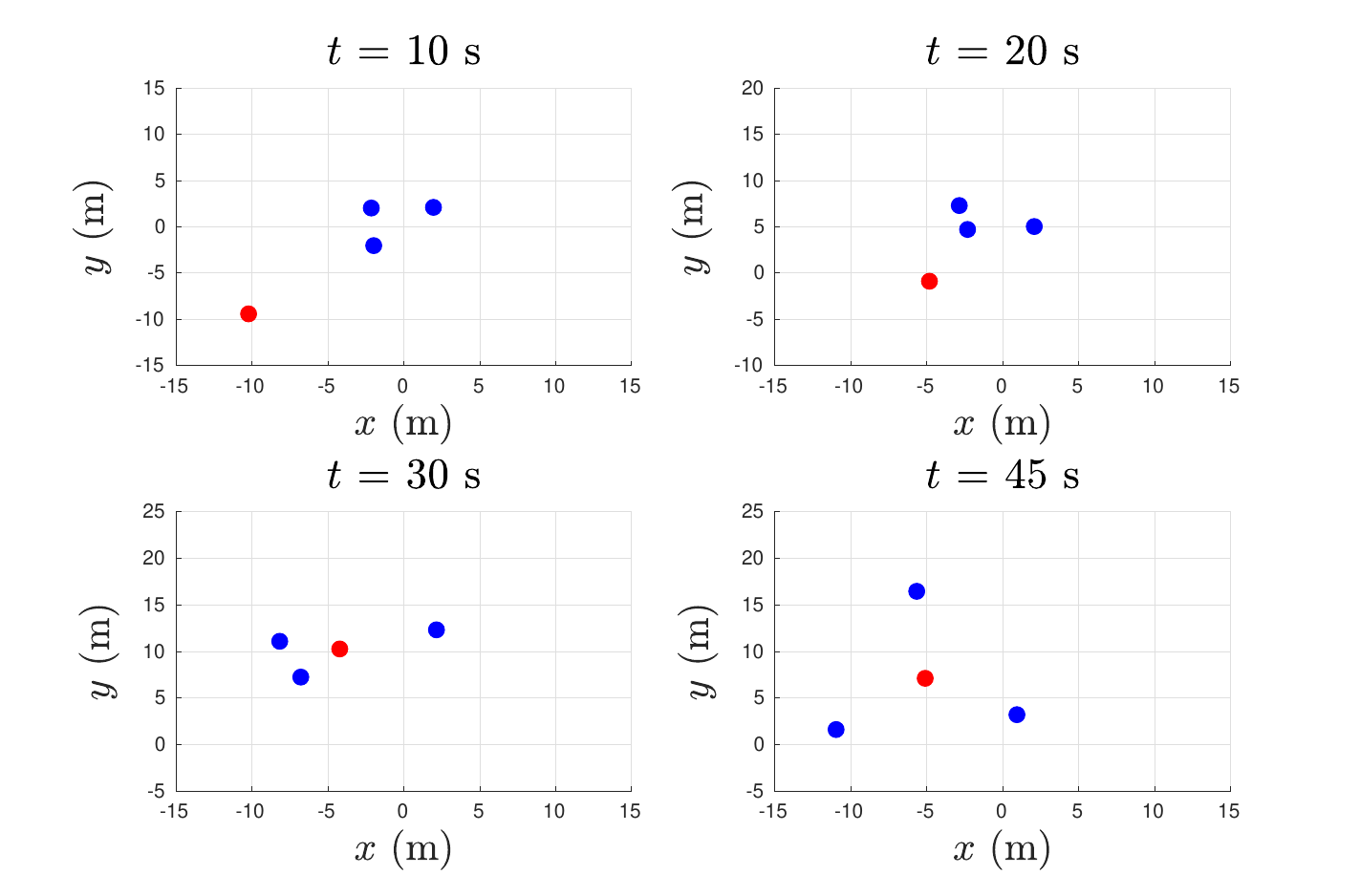}
\caption{Movement of swarm \acp{UAV} (blue) and an interferer (red) in the simulation without using evasive behavior. 
Parameters were enlarged for the stronger swarm reaction to the presence of an detected interferer using the same mechanism as was used for detecting a static obstacle, as proposed in \cite{BioInspiredSwarm}.}
\label{fig:withoutEvasiveBehaviorPositionsEnlargedDetectionRange}
\end{figure}

\begin{figure}[!htb]
\centering
\includegraphics[width=0.85\columnwidth]{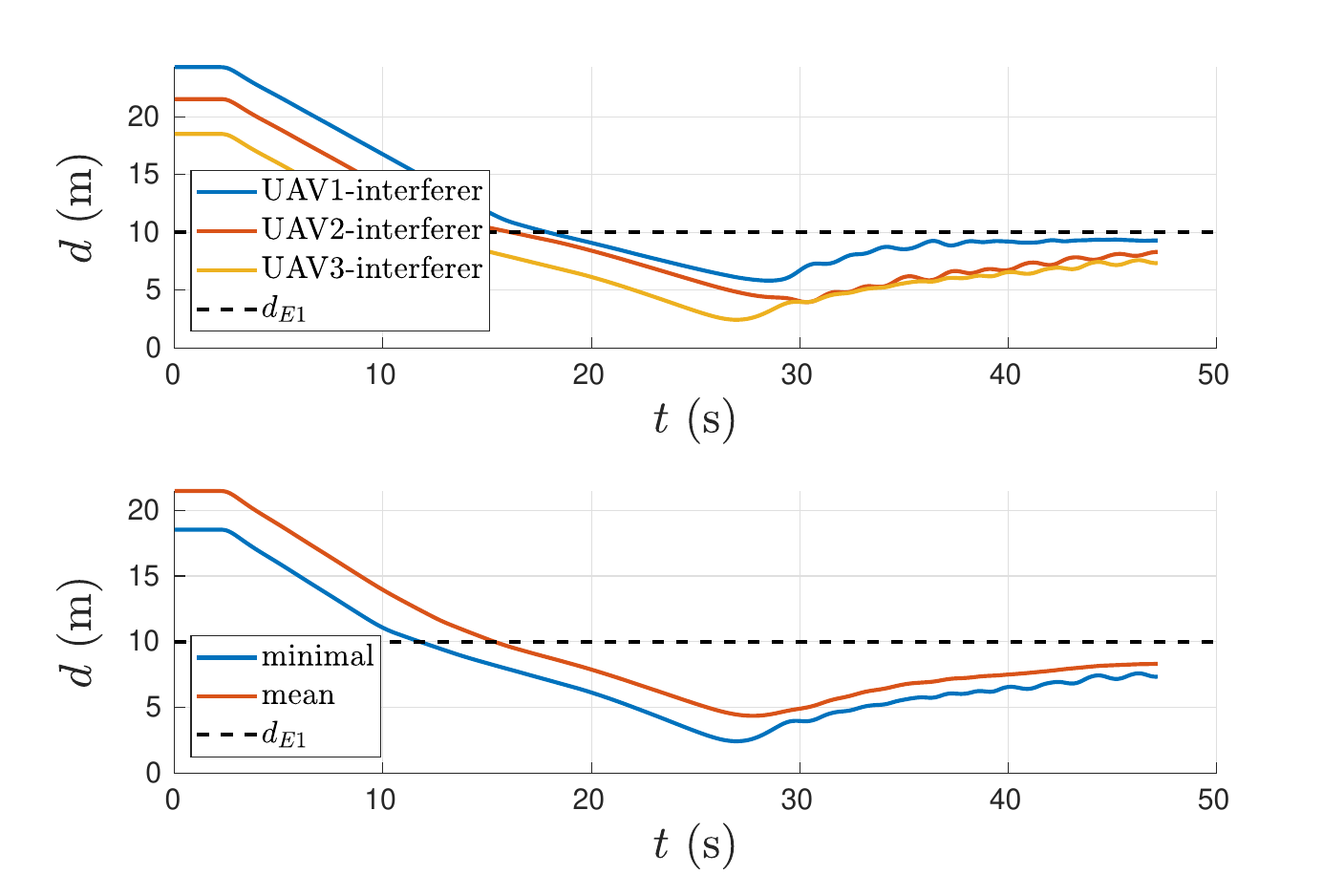}
\caption{The upper figure is a graph of distances between swarm \acp{UAV} and the interferer if all swarm \acp{UAV} do not use evasive behavior. 
Parameters were enlarged for the stronger swarm reaction to the presence of a detected interferer using the same mechanism as was used for detecting a static obstacle, as proposed in \cite{BioInspiredSwarm}. 
The lower figure shows the minimal and mean distance between swarm \acp{UAV} and the interferer.}
\label{fig:withoutEvasiveBehaviorDistPredEnlargedDetectionRange}
\end{figure}

\begin{figure}[!htb]
\centering
\includegraphics[width=0.8\columnwidth]{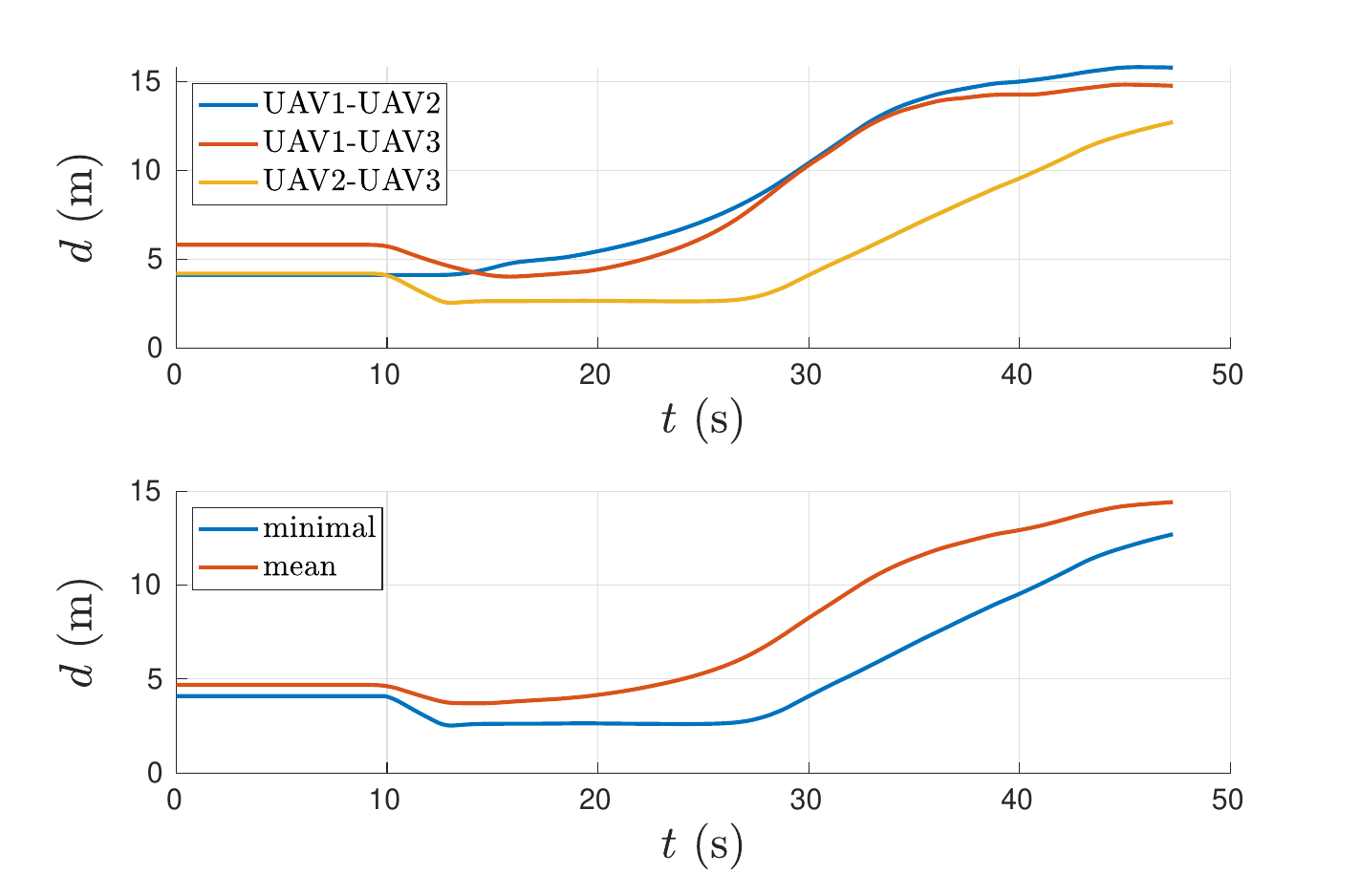}
\caption{The absolute, minimal, and mean distances between swarm \acp{UAV} if all swarm \acp{UAV} do not use evasive behavior. 
Parameters were enlarged for the stronger swarm reaction to the presence of a detected interferer using the same mechanism as was used for detecting a static obstacle, as proposed in \cite{BioInspiredSwarm}.}
\label{fig:withoutEvasiveBehaviorDistUAVsEnlargedDetectionRange}
\end{figure}

This subsection is intended to compare the proposed evasive behavior with state-of-the-art swarming approaches that solve avoidance with dynamic obstacles indirectly by enlarging the detection range \cite{BioInspiredSwarm}. 
In this simulation, the \acp{UAV} can detect an interferer in double the detection range than in the previous simulation.
Moreover, the parameters were enlarged to ensure a stronger reaction to a detected interferer, similar to what is done for a static obstacle.
As shown in Figure~\ref{fig:withoutEvasiveBehaviorPositionsEnlargedDetectionRange}, the movement of swarm \acp{UAV} and an interferer is similar to the previous experiment.
The swarm reacts to the presence of the interferer similar to how it would with a regular swarm member.
There is only one difference: the distances between swarm \acp{UAV} and the interferer are larger (Figure~\ref{fig:withoutEvasiveBehaviorDistPredEnlargedDetectionRange}).
However, the swarm did not escape from the interferer.
The distances between swarm \acp{UAV} shown in Figure \ref{fig:withoutEvasiveBehaviorDistUAVsEnlargedDetectionRange} increased because an interferer divided the swarm into two separate subgroups, as shown in Figure~\ref{fig:withoutEvasiveBehaviorPositionsEnlargedDetectionRange} at $t=30$ s.


\subsection{Simulation -- with evasive behavior}
\label{sec:GazeboEvasiveBehavior}
\begin{figure}[!tb]
\centering
\includegraphics[width=0.8\columnwidth]{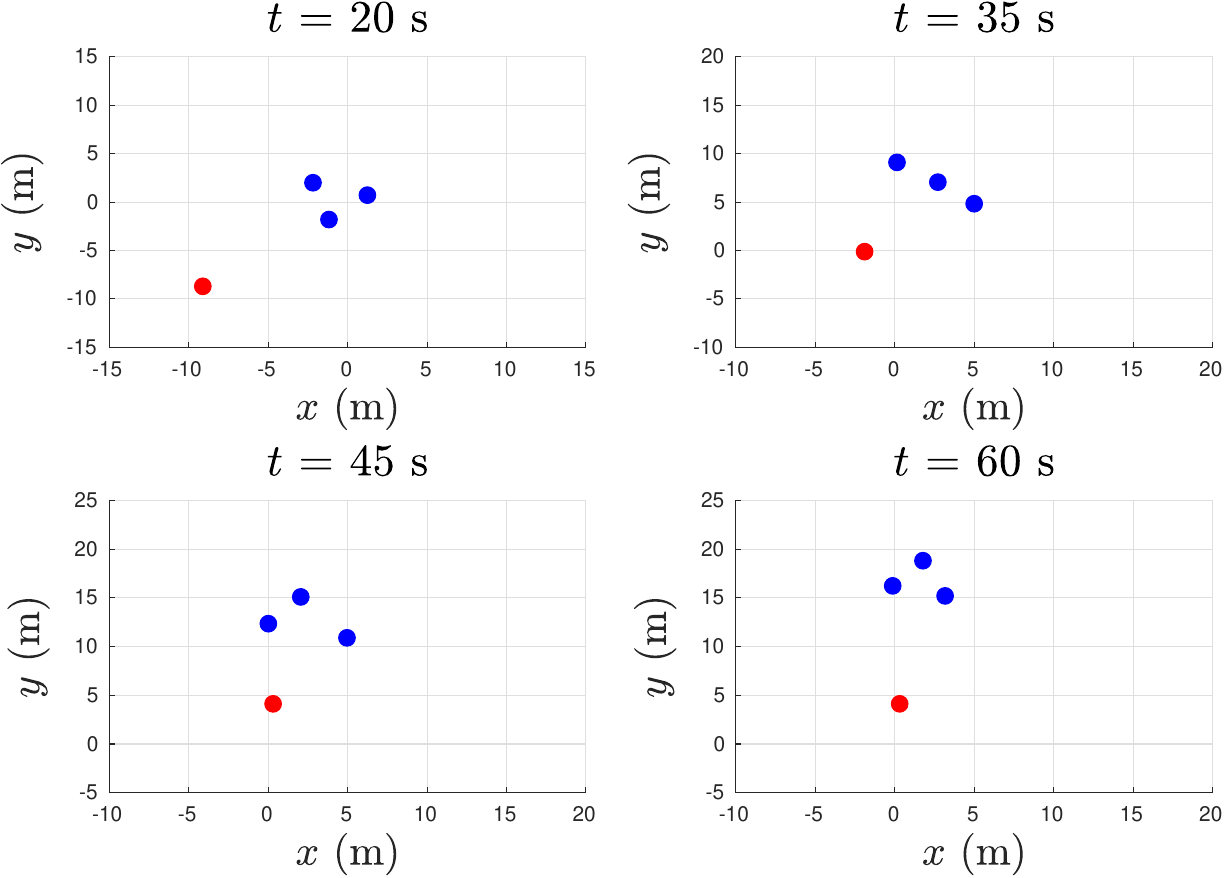}
\caption{Movement of swarm \acp{UAV} (blue) and an interferer (red) in the simulation using evasive behavior.}
\label{fig:withEvasiveBehaviorPositions}
\end{figure}

\begin{figure}[!tb]
\centering
\includegraphics[width=0.865\columnwidth]{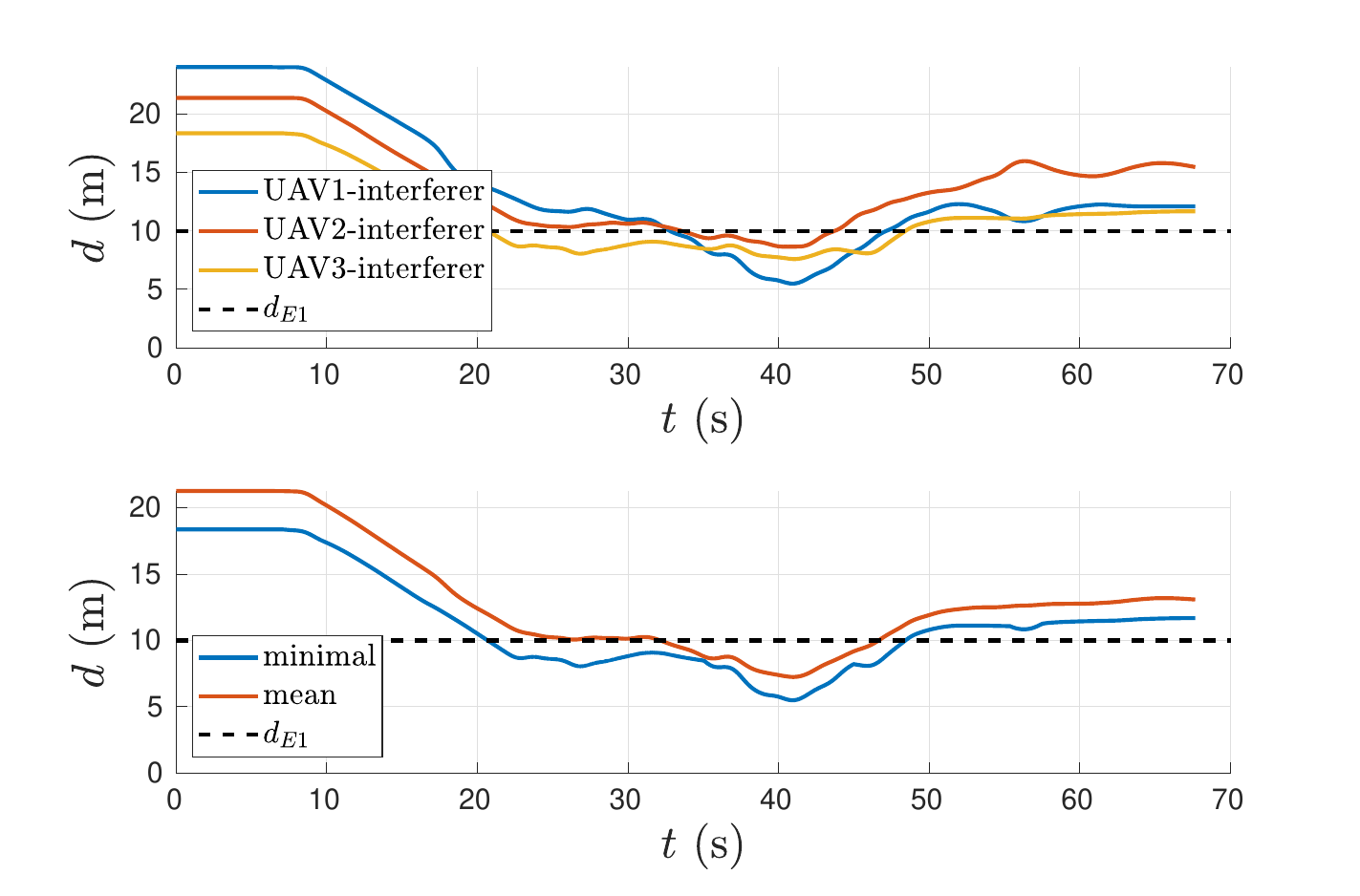}
\caption{The upper figure is a graph of distances between swarm \acp{UAV} and the interferer in the simulation where all swarm \acp{UAV} use evasive behavior. 
The~lower figure shows the minimal and mean distance between swarm \acp{UAV} and the interferer.}
\label{fig:withEvasiveBehaviorDistPred}
\end{figure}

\begin{figure}[!tb]
\centering
\includegraphics[width=0.715\columnwidth]{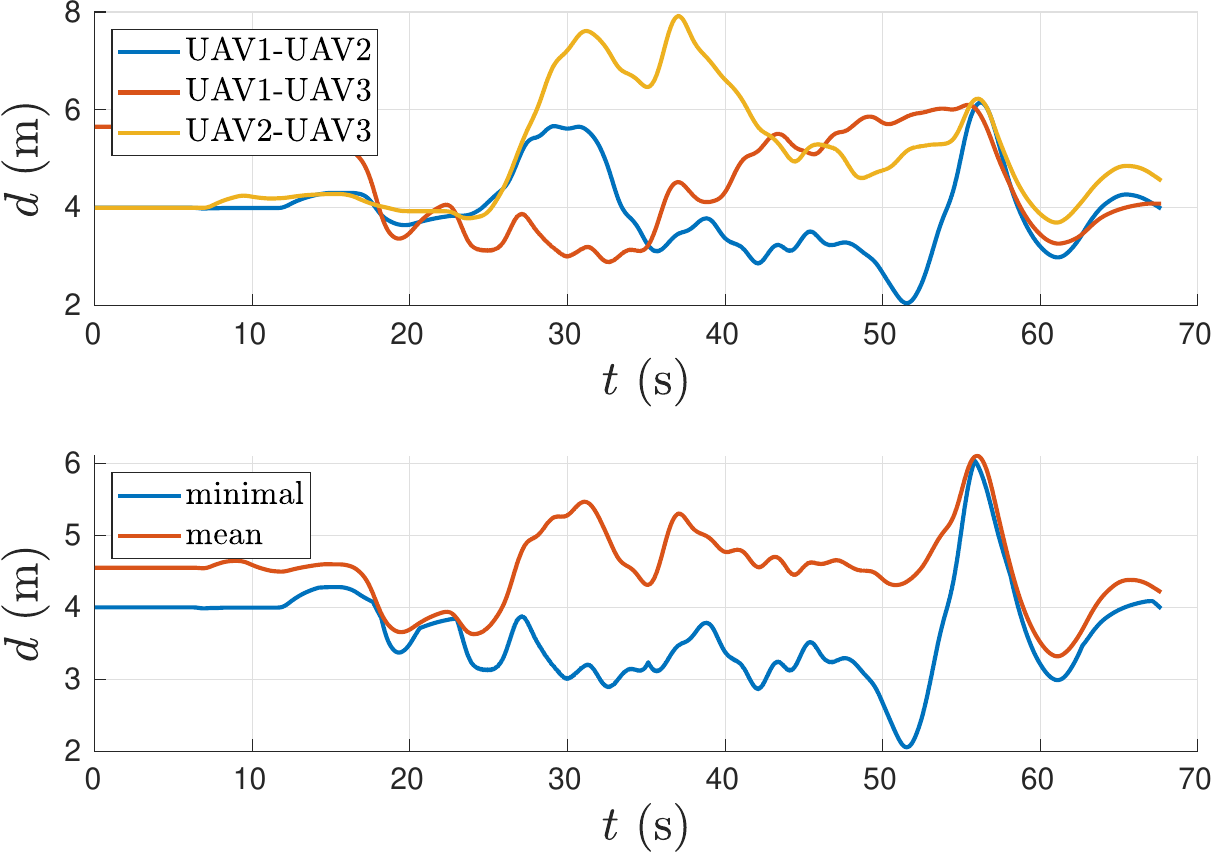}
\caption{The absolute, minimal, and mean distances between swarm \acp{UAV} in the simulation where all swarm \acp{UAV} use evasive behavior.}
\label{fig:withEvasiveBehaviorDistUAVs}
\end{figure}

In the simulation shown in the video available in~\hyperref[footnote-videos]{\footnotemark[\value{footnote}]} and Figures \ref{fig:withEvasiveBehaviorPositions}--\ref{fig:withEvasiveBehaviorDistUAVs}, the swarm uses the proposed evasive behavior.
The initial situation is the same as in the previous experiments, shown in Figure~\ref{fig:4UAVsGazebo} and \ref{fig:withoutEvasiveBehaviorPositionsEnlargedDetectionRange}.
The \ac{UVDAR} system is also used in this experiment for the localization of the interferer and the neighbors by each \ac{UAV}. 
The positions of the swarm \acp{UAV} (blue) and the interferer (red) in $x$, $y$ coordinates at time $t$ are presented in Figure~\ref{fig:withEvasiveBehaviorPositions}.
The figure shows that by using the proposed method, the swarm escapes from the interferer in a compact formation as intended and similar to behavior seen by swarms in nature.
The distances between the swarm \acp{UAV} and the interferer are shown in Figure~\ref{fig:withEvasiveBehaviorDistPred}.
The distance from the swarm to the interferer first slightly diminishes, and then stabilizes above \SI{5}{\meter}. 
Once the interferer stops, the distance of the swarm from the interferer grows again and stabilizes at a distance greater than $d_{E1}$.
The Figure~\ref{fig:withEvasiveBehaviorDistUAVs} shows the mutual distances between the swarm \acp{UAV}. 

To compare the behavior of the system with and without the implemented evasive behavior (section~\ref{sec:GazeboOnlyBoids}), the following time quantities (summarized in Table \ref{table:GazeboSimulationsTimes}) were measured during simulations:
\begin{itemize}
\item $t_s$ is the time when the interferer is at an unsafe distance to any of the swarm members (i.e. the distance to the interferer is less than $d_{E1}$)
\item $t_d$ is the time of the first detection of the interferer by any swarm \ac{UAV},
\item $t_e$ is the time when all swarm \acp{UAV} begin escaping and where all \acp{UAV} are switched to Active or Passive mode.
\item $t_{ed} = t_e - t_d$
\item $t_{ds} = t_d - t_s$
\end{itemize}
\begin{table}[!t]
\centering
\begin{tabular}{|c|c|c|c|c|c|}
\hline
\rowcolor[gray]{0.6}
\makecell{~evasive~\\~behavior~} & $t_s$ [s] & $t_d$ [s] & $t_e$ [s] & $t_{ed}$ [s] & $t_{ds}$ [s]\\
\hline
\makecell{no\\ (sec. \ref{sec:GazeboOnlyBoids})} & 11.352 & 14.268 & X & X & 2.916\\
\hline
\makecell{no\\ (sec. \ref{sec:GazeboOnlyBoidsEnlargedDetectionRange})} & 11.768 & 13.640 & X & X & 1.872\\
\hline
\makecell{yes\\ (sec. \ref{sec:GazeboEvasiveBehavior})} & 20.700 & 21.600 & 21.752 & 0.152 & 0.900 \\
\hline
\end{tabular}
\caption{Quantities $t_s$, $t_d$, $t_e$, $t_{ed}$, and $t_{ds}$ measured in Gazebo simulations presented in sections \ref{sec:GazeboOnlyBoids}, \ref{sec:GazeboOnlyBoidsEnlargedDetectionRange} and \ref{sec:GazeboEvasiveBehavior}.}
\label{table:GazeboSimulationsTimes}
\end{table}
The time $t_{ed}$ shows how fast the entire swarm reacts to the presence of a detected interferer. 
The time $t_{ds}$ shows how fast the interferer was detected when it entered the safe area of any swarm \ac{UAV}.
In a scenario where evasive behavior is not used, the meaning of time $t_{ds}$ is still how fast the entire swarm reacts to the presence of an interferer detected as a static obstacle.
Time $t_{ds}$ when not using evasive behavior (section \ref{sec:GazeboOnlyBoids}) is 20 times greater than time $t_{ed}$ when using evasive behavior (section \ref{sec:GazeboEvasiveBehavior}).
In the case of an enlarged detection range excluding the use of evasive behavior (section \ref{sec:GazeboOnlyBoidsEnlargedDetectionRange}), the time $t_{ds}$ is 12 times greater than time $t_{ed}$ when using evasive behavior (section \ref{sec:GazeboEvasiveBehavior}).
If simulations without the use of evasive behavior are compared, the time $t_{ds}$ is approximately two times smaller if the detection range is enlarged (section \ref{sec:GazeboOnlyBoidsEnlargedDetectionRange}).
The time $t_{ds}$ when using evasive behavior (section~\ref{sec:GazeboEvasiveBehavior}) is 3 times smaller than time $t_{ds}$ without the use of evasive behavior (section \ref{sec:GazeboOnlyBoids}) and 2~times smaller than time $t_{ds}$ without the use of evasive behavior, but with an enlarged detection range (section~\ref{sec:GazeboOnlyBoidsEnlargedDetectionRange}).
Overall, the swarm system reacts much more dynamically and faster when using evasive behavior compared to a situation where evasive behavior is not used.
The distance to the interferer stays larger and the spacing among the individual swarm particles remains safe for flight.
This significantly decreases the probability of mutual collisions in comparison to the basic swarm model~behavior.

\subsection{Experiments with real UAV swarm}
\begin{figure}[!tb]
\centering
\includegraphics[width=0.95\columnwidth]{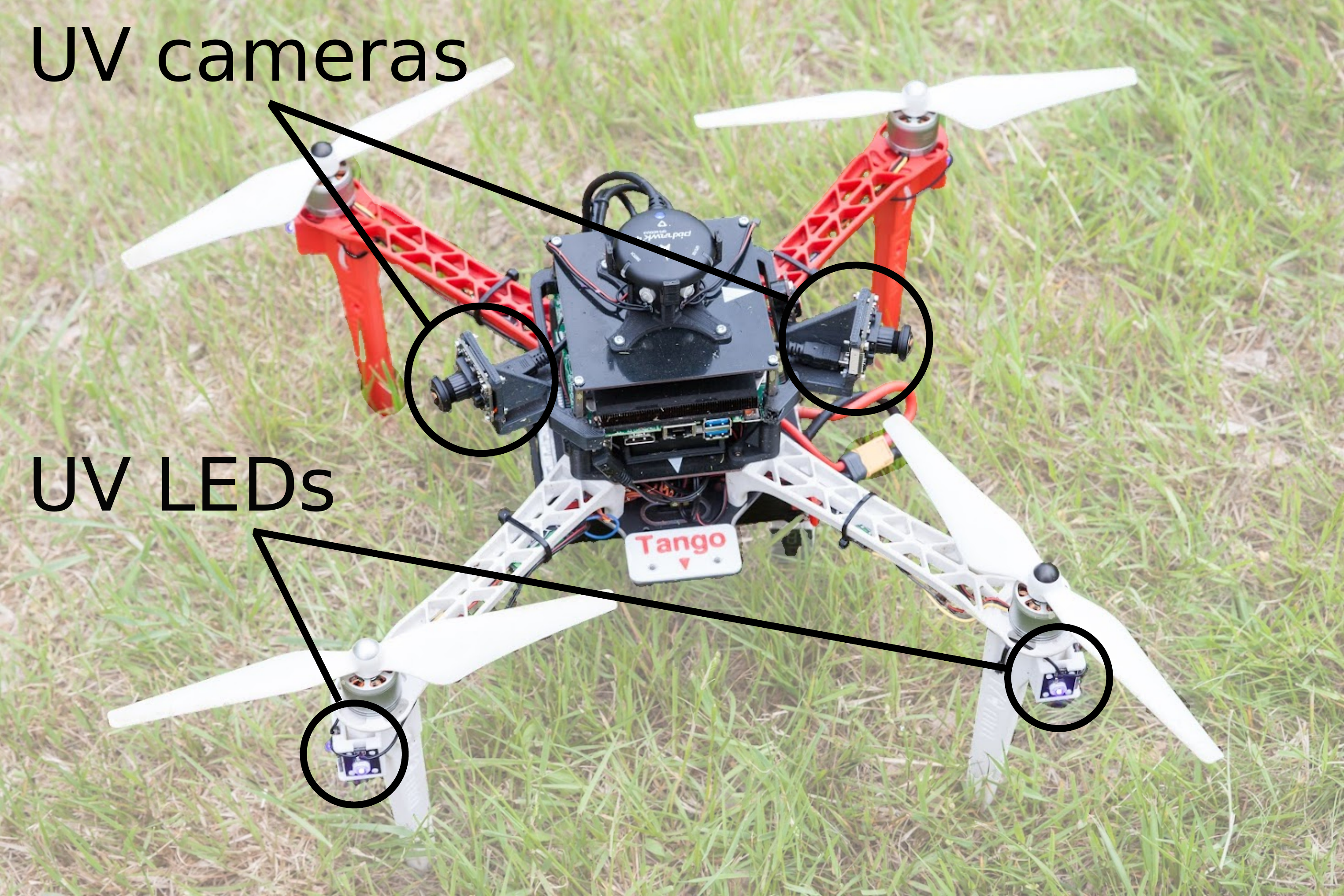}
\caption{UAV used to carry out the experiments equipped with UVDAR system (UV sensitive cameras and UV LEDs).}
\label{fig:realUAV}
\end{figure}

\begin{figure}[!tb]
        \centering
        \subfigure[$t$ = 10 s]{
            \includegraphics[width=0.9\columnwidth]{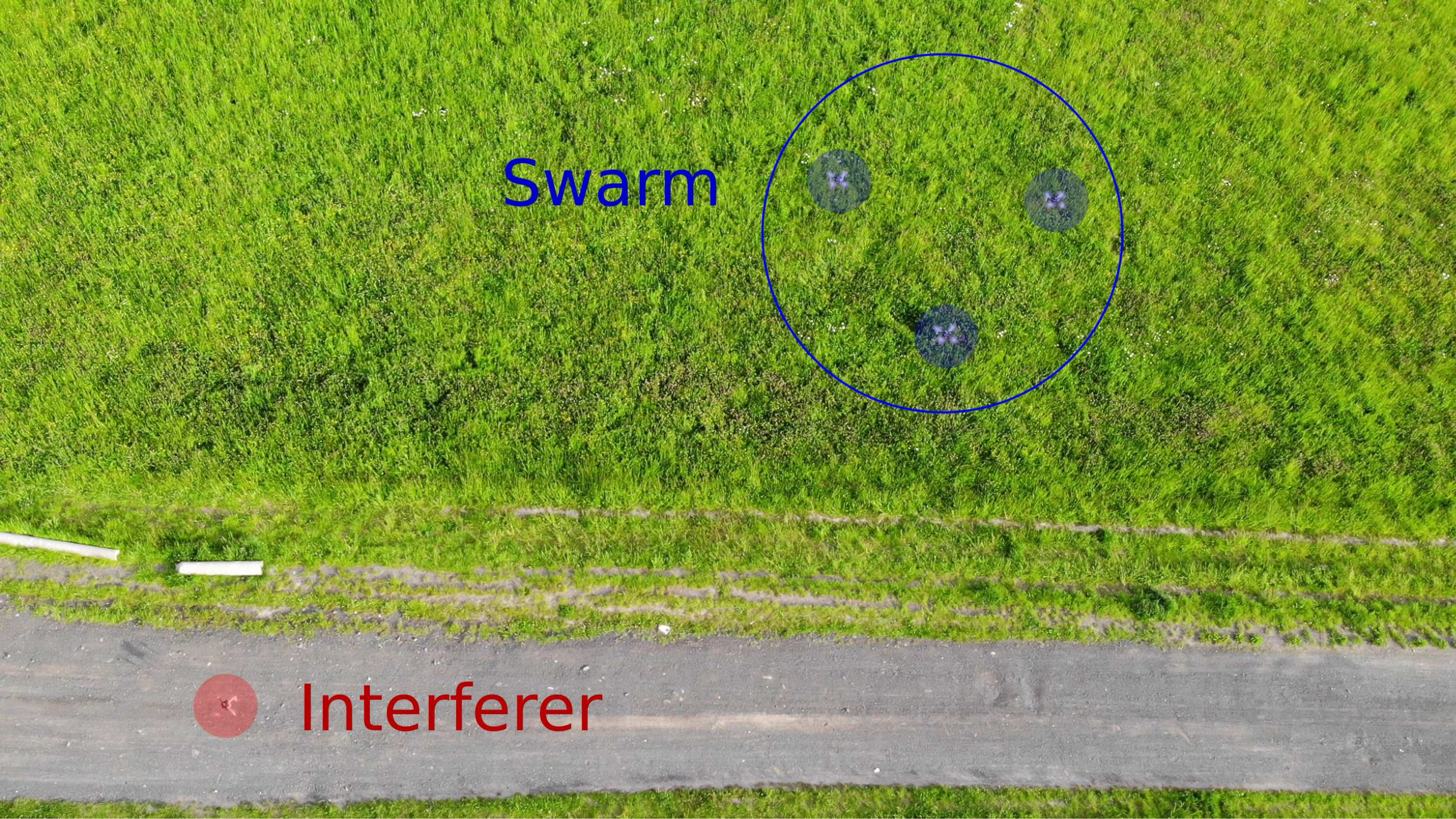}
        }
        \\
         \subfigure[$t$ = 100 s]{   
            \includegraphics[width=0.9\columnwidth]{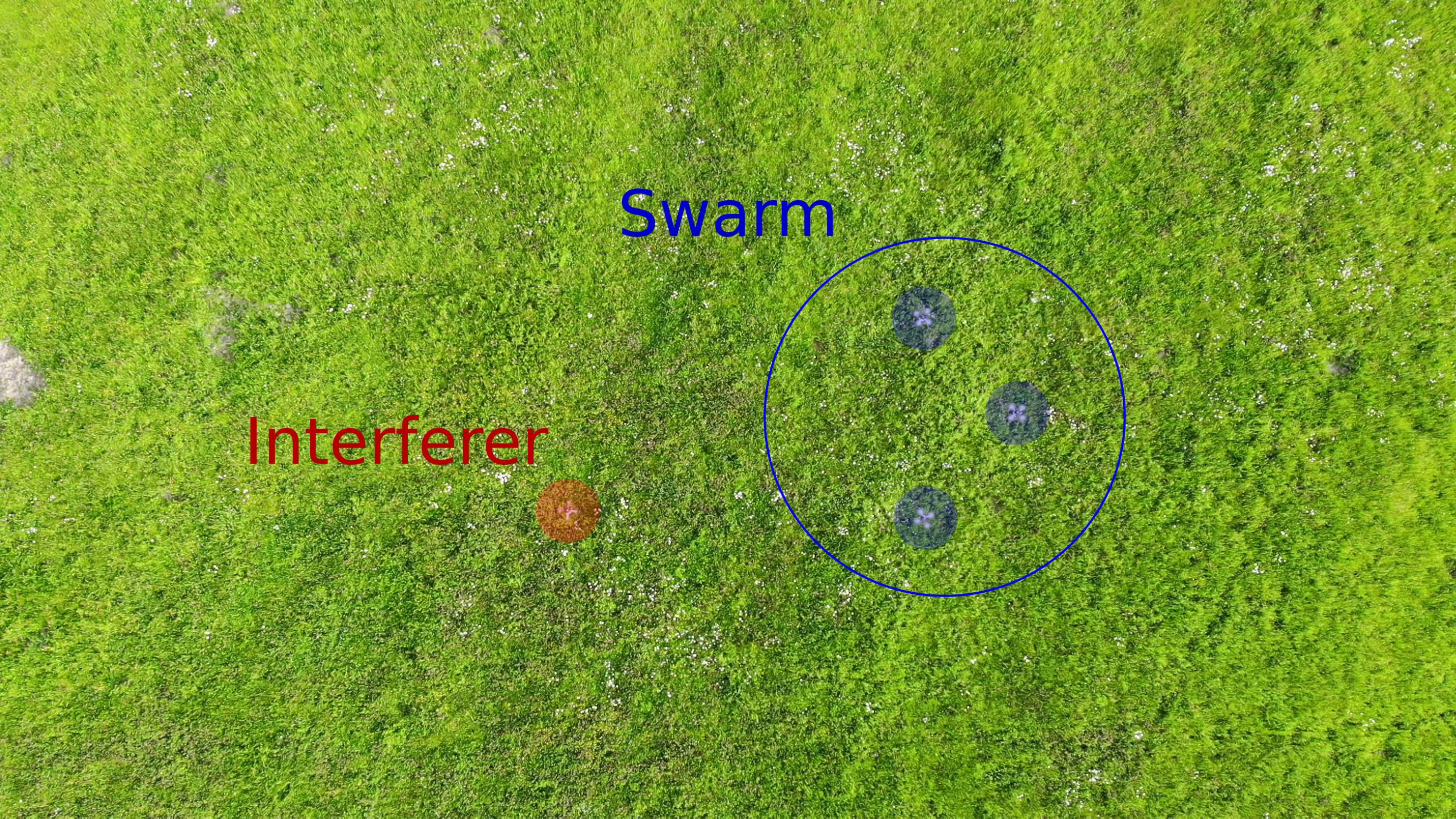}
        }
        \\
		\subfigure[$t$ = 175 s]{  
            \includegraphics[width=0.9\columnwidth]{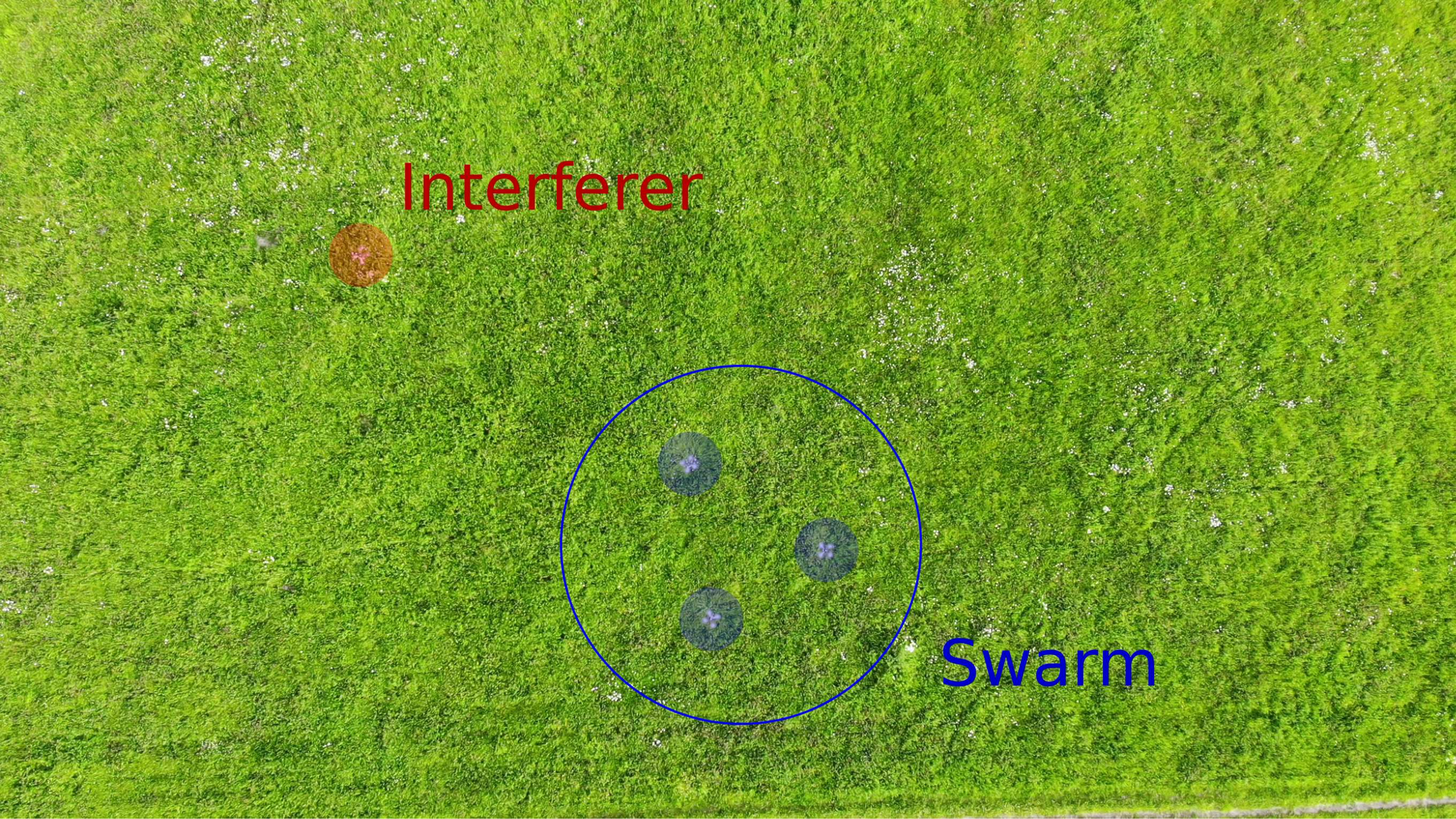}
        }  
        \caption[Snapshots from one of the real-world experiments]{Snapshots from one of the real-world experiments (see video in~\hyperref[footnote-videos]{\footnotemark[\value{footnote}]}).} 
        \label{fig:snapshots}
\end{figure}

\begin{figure}[!tb]
\centering
\includegraphics[width=0.85\columnwidth]{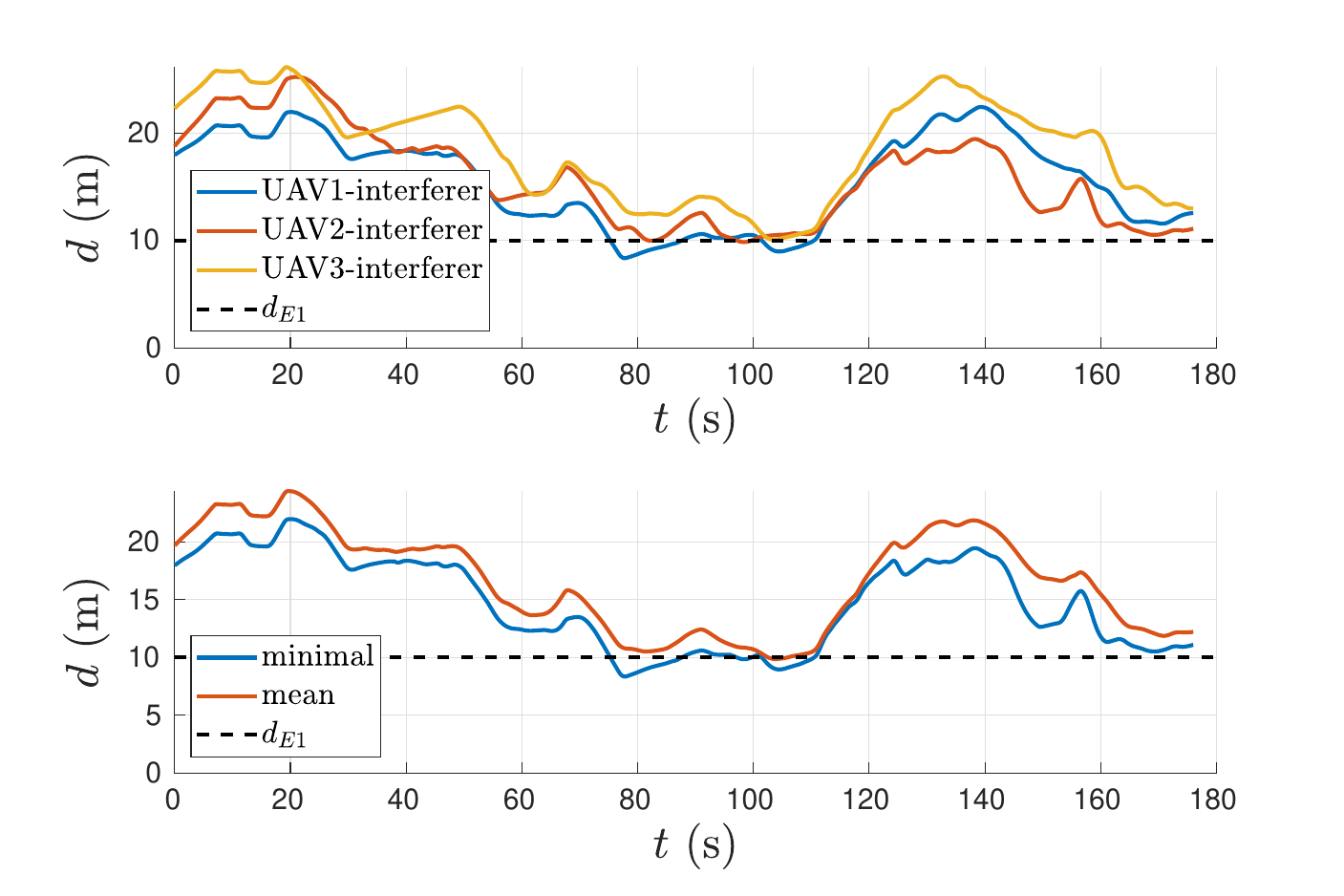}
\caption{The absolute, minimal, and mean distances between swarm \acp{UAV} and the interferer in the real-world experiment in Figure \ref{fig:snapshots}.}
\label{fig:expDistPred}
\end{figure}

\begin{figure}[!tb]
\centering
\includegraphics[width=0.775\columnwidth]{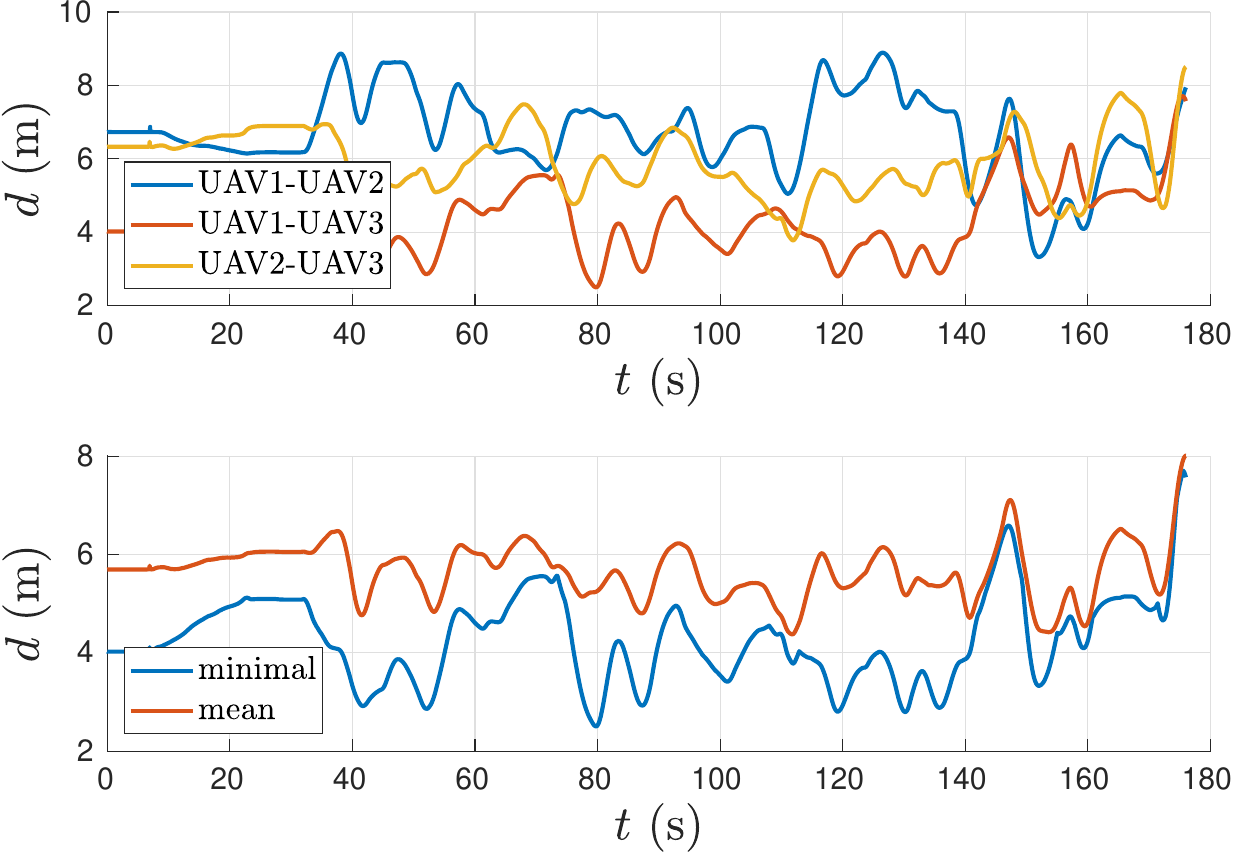}
\caption{The absolute, minimal, and mean distances between swarm \acp{UAV} in the real-world experiment in Figure \ref{fig:snapshots}.}
\label{fig:expDisUAVs}
\end{figure}

After verification and analysis in the Gazebo simulator, the approach proposed in this paper was further verified by conducting real-world experiments in a meadow covering an area of \SI{100}{\meter} by \SI{200}{\meter} (see video in~\hyperref[footnote-videos]{\footnotemark[\value{footnote}]}). 
Four \acp{UAV} were used to carry out these experiments (see Figure \ref{fig:realUAV}), with three composing the swarm and the fourth \ac{UAV} representing the interferer.
At the beginning of the experiments, the swarm \acp{UAV} had been stabilized using the proposed swarm model while the interferer moved towards them. 
After detection of the interferer, the swarm \acp{UAV} began their evasive behavior and were chased by the interferer for 176 s. 
Figures~\ref{fig:snapshots}--\ref{fig:expDisUAVs} present the results of one of the experimental runs, showing that the displayed behavior closely reflects the simulation results.

For mutual \ac{UAV} localization and interferer detection, the \ac{UVDAR} system was used. 
In the real application, an interferer would not be carrying markers known to swarm agents and an onboard computer vision with \ac{CNN} based marker-less \ac{UAV} detection, such as was done in \cite{visionVrba1, visionVrba2}, would be used with the same performance.
The proposed method is designed to be independent to the interferer localization technique.  
The usage of \ac{UVDAR} for this experiment was motivated by the available hardware setup of our experimental platforms carrying only \ac{UV} sensitive cameras and no hardware vision accelerators.
Nevertheless, both relative localization methods - \ac{UVDAR} \cite{UVDAR} and \ac{CNN}-based \cite{visionVrba1} vision - exhibit comparable performance scaling in terms of precision in the case of single robot detection.
To share the position of the detected interferer, the swarm \acp{UAV} need a common global frame to express this information in. 
This was provided by the \ac{GNSS} module. 
Figure~\ref{fig:snapshots} shows snapshots from the real-world experiment, where the position of the swarm and the interferer at different times can be seen. 

The recorded \ac{GNSS} information was used as a ground truth for the quantitative evaluation of the swarm performance in the experiment.
The plot in Figure~\ref{fig:expDistPred} shows the distances between swarm \acp{UAV} and the interferer during the experiment. 
Figure~\ref{fig:expDisUAVs} contains the distances between the swarm \acp{UAV}.
In~the~experiment, the swarm of \acp{UAV} reliably exhibited evasive behavior for the entire duration of \SI{176}{\second}.
The plots of distances in Figures~\ref{fig:expDistPred} and \ref{fig:expDisUAVs} show that no collision had occurred and the minimal distance between the swarm \acp{UAV} was always greater than \SI{2}{\meter}.
This experiment shows that the proposed approach can be applied for reliable swarming of real \acp{UAV} that use onboard relative localization system.

The experimental results with real platforms are comparable to the simulation results, with similar progress of the distances between swarm \acp{UAV} and the interferer in time.
The distances first decrease, then stabilize, and afterwards grow above $d_{E1}$ with the minimal distance in both cases remaining above \SI{2}{\meter}.
Due to noise in real sensors and actuators, the curves of distances between swarm \acp{UAV} fluctuate slightly more in the real-world experiment than in simulation. Still, the proposed approach appeared to be robust enough to achieve the same behavior as was expected from the simulations.

\section{Conclusion}
In this paper, a novel method for the fast collective evasion of self-localized \ac{UAV} swarms has been presented.
The proposed approach may rely on any swarm model adapted for real swarming, such as the one that that was presented in this paper inspired by the Boids model to stabilize a swarm of real \acp{UAV}.
A fast collective evasion approach was proposed to decrease the possibility of collisions with dynamic obstacles or interferers.
Due to the extremely low bandwidth required, the detection of interferers is easily shared among the \acp{UAV} in the swarm using a communication network.
The low bandwidth even enables the use of an implicit visual-based communication integrated into onboard mutual localization.
In context of the shock propagation that occurs in natural swarms, the spread of information that an interferer was detected was simulated using a simplified swarm physics model to show that the method can be effective, even for swarms with a large number of agents. 

To imitate more realistic conditions, the proposed approach has been tested in a realistic simulator, as well as in real world flights.
In these experiments, the proposed approach utilizing the implemented method for evasive behavior was compared with a system using only a basic swarm model.
This comparison has shown the importance of cooperative evasive behavior in addressing interferers, as without it the hazard was not sufficiently addressed by the swarm model alone.
With the evasive behavior active, the swarm of real \acp{UAV} was able to repeatedly escape from an interferer at a safe distance without any collision among the swarm members.

\section*{Acknowledgments}
The presented work has been supported by CTU grant no SGS20/174/OHK3/3T/13, by the Czech Science Foundation (GAČR) under research project No. 20-10280S and by OP VVV funded project CZ.02.1.01/0.0/0.0/16 019/0000765 "Research Center for Informatics".
Our thanks also belong to Daniel He\v{r}t who prepared the necessary equipment for real-world experiments.

\section*{Supplementary materials}
The multimedia materials supporting the article are available at \url{http://mrs.felk.cvut.cz/papers/uav-swarm-fast-collective-evasion}.
The entire UAV system is also available as open-source at \url{https://github.com/ctu-mrs}.

\section*{ORCID}
Filip Nov\'{a}k $\orcidicon{0000-0003-3826-5904}$: 0000-0003-3826-5904\\[0.1cm]
Viktor Walter $\orcidicon{0000-0001-8693-6261}$: 0000-0001-8693-6261\\[0.1cm]
Pavel Petr\'{a}\v{c}ek $\orcidicon{0000-0002-0887-9430}$: 0000-0002-0887-9430\\[0.1cm]
Tom\'{a}\v{s} B\'{a}\v{c}a $\orcidicon{0000-0001-9649-8277}$: 0000-0001-9649-8277\\[0.1cm]
Martin Saska $\orcidicon{0000-0001-7106-3816}$: 0000-0001-7106-3816 

\section*{References}
\bibliographystyle{IEEEtran}
\bibliography{main}

\end{document}